\documentclass[11pt]{article}

\usepackage[preprint]{acl}

\usepackage{times}
\usepackage{latexsym}
\usepackage{amsmath}
\usepackage{amssymb}
\usepackage{booktabs}
\usepackage{multirow}
\usepackage[table]{xcolor}

\usepackage[T1]{fontenc}
\usepackage[utf8]{inputenc}

\usepackage{microtype}

\usepackage{inconsolata}

\usepackage{graphicx}

\title{STAMP: Training Explicit Memory for Mobile GUI Agents in Controllable and Scalable Virtual Environments}

\author{
 \textbf{Junyang Wang\textsuperscript{2,1}\thanks{Work done during internship at Tongyi AI Lab.}},
 \textbf{Haiyang Xu\textsuperscript{1}\thanks{Corresponding author}},
 \textbf{Xi Zhang\textsuperscript{1}},
 \textbf{Zhaoqing Zhu\textsuperscript{1}},
\\
 \textbf{Ming Yan\textsuperscript{1}\footnotemark[2]},
 \textbf{Jieping Ye\textsuperscript{1}},
 \textbf{Jitao Sang\textsuperscript{2}\footnotemark[2]},
\\
 \textsuperscript{1}Tongyi AI Lab, Alibaba Group
 \textsuperscript{2}Beijing Jiaotong University
\\
 \small{
   {\{junyangwang, jtsang\}@bjtu.edu.cn}
 }
\\
 \small{
   {\{shuofeng.xhy, ym119608\}@alibaba-inc.com}
 }
}

\begin{document}
\maketitle
\begin{abstract}
Mobile GUI agents excel at immediate reactive control but frequently fail in realistic, long-horizon tasks that require memory. This failure stems from a fundamental conflict between limited context windows and token-heavy screenshots. To save the limited context, agents must progressively discard older visual history, permanently losing crucial transient information. Furthermore, existing action-centric datasets fail to teach agents what or when to explicitly memorize, and augmenting static real-world data is prohibitively expensive and lacks interactive verification. To resolve this, we present STAMP, a framework that trains explicit memory in mobile agents through controllable virtual environments, where deterministic memory variables are programmatically injected into synthesized tasks to control what must be memorized, when it should be encoded, and when it must later be retrieved, thereby producing verifiable supervised data at scale and enabling online reinforcement learning through environment-driven reward feedback. Evaluated on our newly introduced Memory-World benchmark, the resulting Stamp-GUI agent achieves state-of-the-art performance among GUI-specialized models and sets a new high watermark on our Memory-World benchmark, demonstrating exceptional memory accuracy and task resilience while maintaining strong general mobile navigation capabilities.
\end{abstract}

\section{Introduction}

Mobile GUI agents aim to assist users by directly perceiving and interacting with smartphone interfaces~\citep{qin2025ui,xu2026mobile,bai2025qwen3,zhou2025mai,team2026ui,wang2024mobile,wang2024mobile2,wang2025mobile,ye2025mobile,zhang2025appagent,yan2025step,liu2024autoglm,Agent-S2,gu2025ui,wang2025ui}. While existing end-to-end agents excel in benchmarks focused on immediate perception and reactive control, they frequently fail in realistic, long-horizon scenarios such as cross-platform price comparison~\citep{rawles2025androidworld,li2025screenspot,kong2025mobileworld,chen2026knowu}. In these practical settings, the performance bottleneck is not poor user interface perception, but a fundamental deficit in memory: the capacity to retain and precisely retrieve transient, task-specific information across multiple interaction steps.

This memory failure stems from a fundamental conflict between limited context windows and token-heavy images~\citep{zhang2024large,nguyen2024gui,wang2024gui,laurenccon2024matters}. Because high-resolution screenshots are computationally expensive, it is impossible to feed an agent's entire visual history into a single context window. To solve this, typical approaches compress or truncate the interaction history by progressively discarding older screenshots. To maintain basic task progress, agents often rely on auxiliary text streams, such as maintaining a running dialogue history~\citep{qin2025ui,zhou2025mai} or generating abstract action summaries~\citep{bai2025qwen3,xu2026mobile}.

\begin{figure}[t]
    \centering
    \includegraphics[width=1\linewidth]{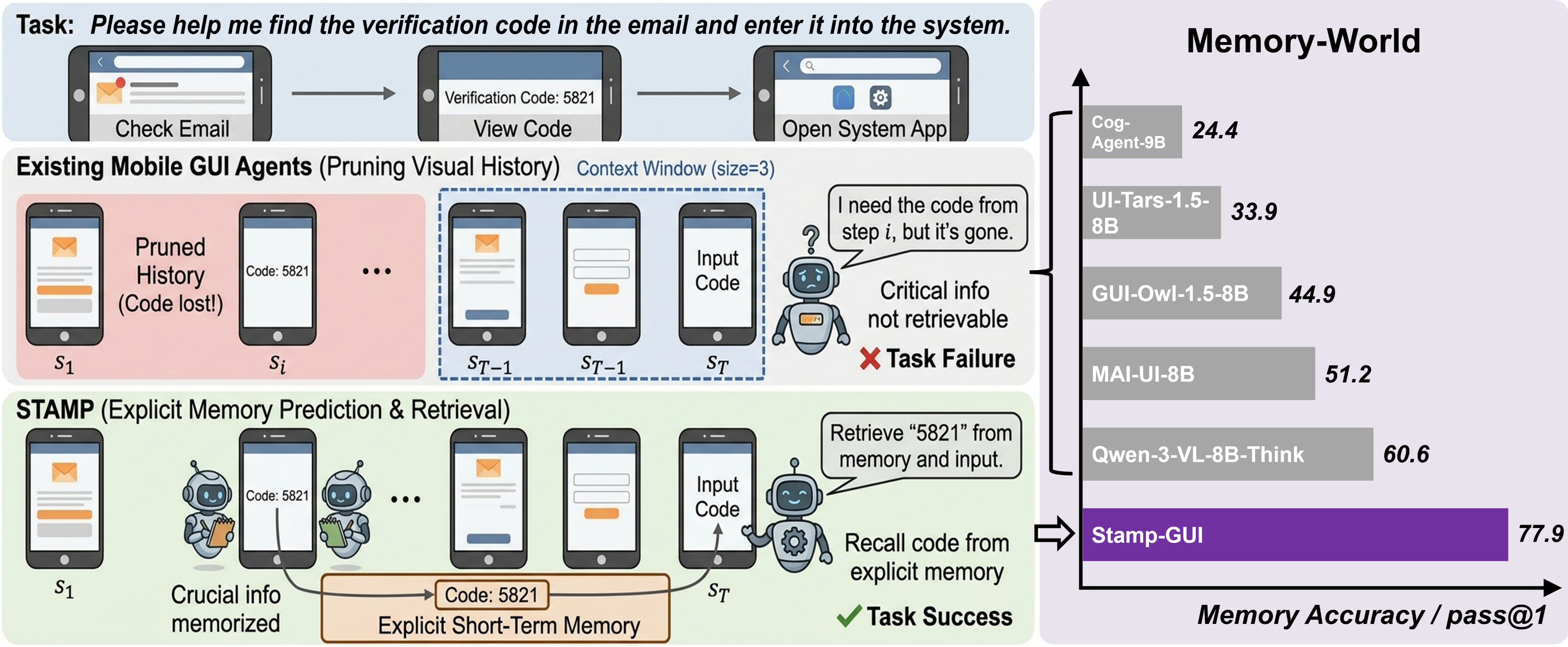}
    \caption{Existing agents fail by pruning critical information as history moves out of the context window, whereas Stamp-GUI can memorize and retrieve crucial information when needed.}
    \label{fig:intro}
    \vspace{-3mm}
\end{figure}

However, as shown in Figure~\ref{fig:intro}, these text-centric compression methods fail to solve the root problem. Since throwing away historical images is an unavoidable architectural necessity, any visual detail not explicitly captured in the text summary—such as a brief verification code or a specific item price—is permanently lost. When a downstream step eventually requires this vanished information, the agent falls into an irrecoverable blind spot, inevitably leading to hallucinations or task failure.

To prevent this information loss, an agent must actively predict and output critical memories into its text context before the source image is discarded. Unfortunately, existing mobile GUI datasets are overwhelmingly action-centric, optimizing only for physical operations~\citep{rawles2023androidinthewild,zhang2024android,li2024effects,lu2025guiodyssey,chai2025amex,luo2025gui,lu2026ui,cheng2024seeclick}. As a result, models never learn what specific information is worth remembering or when to record it. As shown in Figure 1, agents trained purely on traditional data achieve very poor Memory Accuracy on our Memory-World benchmark. A straightforward idea is to augment existing real-world datasets with explicit memory labels. However, this is highly challenging: using proprietary LLMs for post-hoc labeling of real-world trajectories is expensive, noisy, and struggles to pinpoint the exact moment information should be memorized. More importantly, static real-world datasets cannot provide interactive feedback to verify if the memorized content is factually accurate.

To overcome these data limitations, we present STAMP, a fully automated framework that trains explicit memory in mobile agents via controllable and scalable virtual environments. Instead of relying on static real-world data, STAMP programmatically synthesizes large numbers of memory-intensive tasks with diverse layouts, semantics, and interaction flows. Crucially, in these virtual sandboxes, we can explicitly control what transient information should be memorized, when it appears, and at which later step it must be retrieved for successful completion. This controllability makes the supervision automatically verifiable at both the task and memory levels, while the simulator itself provides an efficient path to scale data generation and online exploration. As a result, STAMP not only produces high-quality supervised fine-tuning data but also enables large-scale online reinforcement learning with environment-driven reward feedback. By exploring alternative execution paths in the simulator and receiving immediate reward feedback on task success, the agent explicitly learns to optimize its memory-encoding actions. Using this framework, we train Stamp-GUI and introduce Memory-World, a benchmark designed specifically to isolate and quantify memory performance in mobile interfaces.

Our contributions are summarized as follows:
\begin{itemize}
\item We identify the inevitable visual information loss in mobile agents and propose STAMP, a framework that leverages controllable and scalable virtual environments to automatically generate verifiable explicit-memory data.
\item We introduce Stamp-GUI, an end-to-end agent trained with both supervised fine-tuning and online reinforcement learning on virtual environments where memory targets are precisely controlled and automatically verified.
\item We provide Memory-World along with memory-focused adaptations of existing benchmarks to facilitate the rigorous evaluation of memory capabilities in mobile agents.
\end{itemize}

\begin{figure*}[!ht]
    \centering
    \includegraphics[width=1\linewidth]{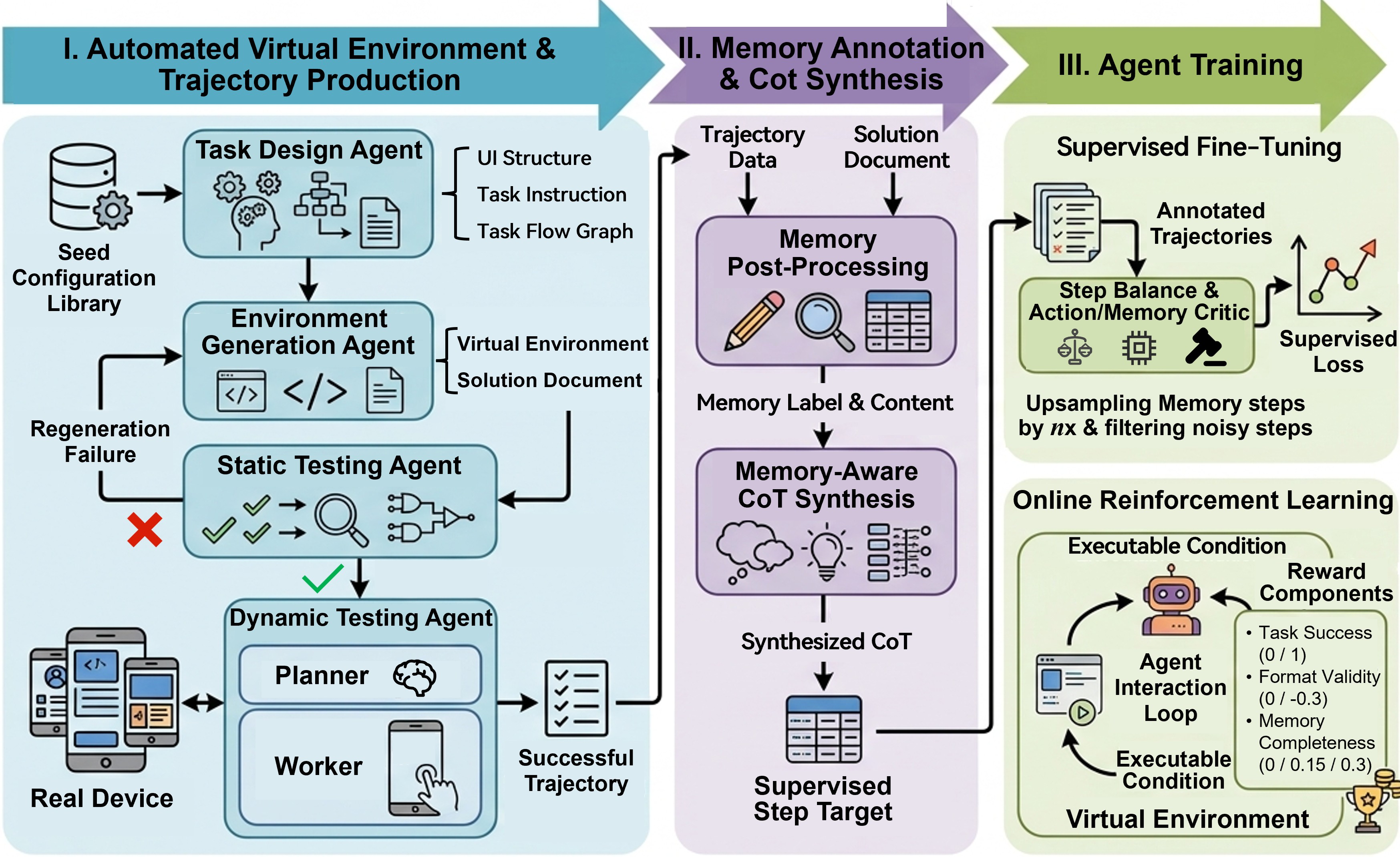}
    \caption{The end-to-end framework of STAMP. The proposed pipeline consists of three sequential phases: 
    \textbf{Phase I (Left):} Seed configuration guided automated virtual environment synthesis and dynamic expert trajectory collection. 
    \textbf{Phase II (Middle):} Successful trajectory post-processing, memory data isolation, and step-level CoT target annotation. 
    \textbf{Phase III (Right):} Model training via offline step-balanced supervised fine-tuning and online sandbox reinforcement learning.}
    \label{fig:framework}
\end{figure*}

\section{Related Works}

\subsection{Mobile GUI Agent}
The field of Mobile GUI agents has transitioned from early rule-based systems and those dependent on structured metadata, such as accessibility trees~\citep{rawles2025androidworld}, to native multimodal vision-language models that perceive raw screenshots directly~\citep{qin2025ui, xu2026mobile}. Modern autonomous agents, such as the GUI-Owl series, leverage large-scale pre-training and unified action spaces to achieve human-like interactions across diverse platforms. Recent advancements have introduced complex reasoning architectures, such as the "Thinking" mode in GUI-Owl-1.5~\citep{xu2026mobile} and the device-cloud collaborative system in MAI-UI~\citep{zhou2025mai}, which significantly improve task success rates in long-horizon mobile navigation.

\subsection{Memory in GUI Agent}
Currently, most methods handle memory only indirectly, either by relying on long dialogue contexts that encompass entire interaction histories~\citep{qin2025ui, zhou2025mai} or through compressed textual summaries of previous screens~\citep{bai2025qwen3, xu2026mobile}. These approaches are susceptible to information dilution or omission as the task complexity grows, leading to catastrophic failures in memory-intensive tasks such as cross-app verification or price comparison~\citep{liu2026memgui, xu2026mobile}. Our work addresses this limitation by proposing STAMP, a framework that explicitly trains mobile agents for memory using controllable virtual environments. Unlike prior indirect methods, our approach empowers the agent to proactively predict and store critical information, establishing memory as an explicit component of the action prediction loop.

\section{Method}

\subsection{Overview}

We propose STAMP, a framework for training explicit memory in mobile GUI agents via controllable virtual environments. The key idea is to synthesize tasks in which transient information appears at one step, disappears from the visual context after subsequent interactions, and must be retrieved later for successful completion. This enables automatic construction of verifiable supervision for both memory prediction and task execution.

\subsection{Problem Setup}

A mobile task is defined as $x=(q,e)$, where $q$ is the user instruction and $e$ is an executable environment. A trajectory is
\[
\tau=\{(s_t,y_t,s_{t+1})\}_{t=1}^{T},
\]
where $s_t$ is the state and $y_t$ is the model output at step $t$. In Stamp-GUI, the model output is
\[
y_t=(c_t,a_t,m_t),
\]
where $c_t$ denotes the reasoning content, $a_t$ the action, and $m_t$ the memory content, with $m_t=\varnothing$ when no memorization is required. We therefore model
\[
p_{\theta}(c_t,a_t,m_t \mid q,s_{\le t},y_{<t}).
\]

Unlike standard GUI agents that only predict actions, Stamp-GUI is trained to proactively externalize critical transient information into $m_t$ before the source screenshot is dropped from the context window.

\subsection{Virtual Environment Data Construction}

STAMP uses controllable virtual mobile environments to synthesize memory-intensive tasks at scale. Each environment is generated from a sampled seed configuration controlling latent factors such as style, layout density, and semantic content. We explicitly inject memory dependencies into each task by specifying what information must be memorized, when it appears, and at which later step it must be retrieved.

After static verification, we execute the generated environments to collect successful trajectories. Because the memory-bearing variables are programmatically controlled, each successful trajectory can be aligned with the environment specification to recover step-level memory supervision:
\[
\mathcal{M}(\tau)=\{(b_t,\tilde{m}_t)\}_{t=1}^{T}, \qquad b_t\in\{0,1\},
\]
where $b_t$ indicates whether step $t$ requires memorization and $\tilde{m}_t$ is the corresponding target memory, with $\tilde{m}_t=\varnothing$ when $b_t=0$. This gives an augmented trajectory
\[
\tilde{\tau}=\{(s_t,\hat{a}_t,\tilde{m}_t,s_{t+1})\}_{t=1}^{T}.
\]

We then synthesize memory-aware reasoning targets for each step and apply a step-level action/memory critic to filter noisy labels before training.

\subsection{Supervised Training}

The final supervised target at each step is
\[
\tilde{y}_t=(\tilde{c}_t,\hat{a}_t,\tilde{m}_t),
\]
where $\tilde{c}_t$ is synthesized reasoning conditioned on the state, action, and memory target. Since memory-required steps are much sparser than ordinary action steps, we upweight them during supervised fine-tuning. Specifically, we use $w_t^{\mathrm{bal}}=n$ when $b_t=1$ and $1$ otherwise, and optimize
\[
\begin{aligned}
\mathcal{L}_{\mathrm{SFT}}
&= \sum_{t=1}^{T} w_t^{\mathrm{bal}} (
    w_t^a \mathcal{L}_{a}(a_t,\hat{a}_t) \\
&\hphantom{= \sum_{t=1}^{T} w_t^{\mathrm{bal}} (}
    {}+ w_t^m \mathcal{L}_{m}(m_t,\tilde{m}_t)
    + \mathcal{L}_{c}(c_t,\tilde{c}_t)
).
\end{aligned}
\]

where invalid action or memory labels identified by the critic are masked out during training.

\subsection{Online Reinforcement Learning}

We further refine the model with online reinforcement learning in executable virtual environments. Since each environment exposes both executable success conditions and injected memory targets, task completion and memory behavior are directly rewardable. For a rollout trajectory $\tau$, the total reward is
\[
r(\tau)=r_{\mathrm{task}}(\tau)+r_{\mathrm{fmt}}(\tau)+r_{\mathrm{mem}}(\tau).
\]
The task reward is
\[
r_{\mathrm{task}}(\tau)=\mathbb{I}[\,\tau \text{ succeeds}\,],
\]
and the format reward is
\[
r_{\mathrm{fmt}}(\tau)=
\begin{cases}
0, & \text{if the output format is valid},\\
-0.3, & \text{otherwise}.
\end{cases}
\]

For memory reward, we compare the model-generated memory trace over the full rollout against the reference memory trace extracted from post-processing. Let
\[
M^{\star}(\tau)=\{\tilde{m}_t \mid b_t=1\},
\hat{M}(\tau)=\{\hat{m}_t \mid b_t=1\},
\]
denote the reference and predicted memory contents over all memory-required steps in $\tau$. A judge model evaluates the trajectory-level match between $\hat{M}(\tau)$ and $M^{\star}(\tau)$ and returns a completeness score
\[
\Gamma(\tau)\in\{1,0.5,0\},
\]
corresponding to complete, partial, and insufficient memory coverage, respectively. We then define
\[
r_{\mathrm{mem}}(\tau)=0.3 \cdot \Gamma(\tau).
\]

This trajectory-level formulation is important because many tasks contain multiple memory events. Rewarding only the best individual memory step would overestimate memory quality, whereas our design encourages faithful memory prediction throughout the entire interaction.

\section{Experiments}

\subsection{Benchmarks and Metrics}
We evaluate on three diverse benchmarks focusing on mobile GUI interactions and memory capabilities: AndroidWorld-M, MemGUI-Bench, and our proposed Memory-World. For all benchmarks, we report \textit{pass@1} and \textit{pass@3} to measure single-attempt and multi-attempt success rates. 

For both AndroidWorld-M and MemGUI-Bench, we use the same three-level difficulty protocol to better separate memory from general action difficulty: \textbf{L1} simplifies the task configuration and provides task-required knowledge, \textbf{L2} keeps the original configuration while providing task-required knowledge, and \textbf{L3} uses the original task setting without additional assistance.

For Memory-World, alongside the standard Task Success Accuracy (\textbf{T-Acc}), we introduce Memory Accuracy (\textbf{M-Acc}) to directly measure the fidelity of the agent's generated memory against the ground-truth transient knowledge:
\[
\mathrm{M\mbox{-}Acc} = \frac{1}{N}\sum_{i=1}^{N} g_i, \quad g_i \in \{1, 0.5, 0\}
\]
where $g_i$ represents the semantic matching score scored by an automated evaluator. In particular, Memory-World is built from controllable virtual environments, allowing us to isolate memory variables and automatically verify both task outcomes and memory correctness.

\begin{table*}[t]
\centering
\renewcommand{\arraystretch}{0.9}
\small
\setlength{\tabcolsep}{9pt}
\begin{tabular}{l|ccc>{\columncolor{gray!15}}c|ccc>{\columncolor{gray!15}}c}
\toprule
\multirow{2}{*}{\textbf{Model}} & \multicolumn{4}{c}{\textbf{\textit{pass@1}}} & \multicolumn{4}{c}{\textbf{\textit{pass@3}}} \\
\cmidrule(lr){2-5} \cmidrule(lr){6-9}
 & L1 & L2 & L3 & Avg. & L1 & L2 & L3 & Avg. \\
\midrule
\multicolumn{9}{l}{\textit{General-purpose Multimodal Model + Active Memory Instruction (AMI)}} \\
\midrule
Qwen3-VL-8B-Instruct \citep{bai2025qwen3} & 15.4 & 7.7 & 0.0 & 7.7 & 23.1 & 15.4 & 7.7 & 15.4 \\
Qwen3-VL-8B-Think \citep{bai2025qwen3} & 23.1 & 15.4 & 0.0 & 12.8 & 30.8 & 23.1 & 7.7 & 20.5 \\
GPT-5.4 \citep{gpt54} & 23.1 & 15.4 & 7.7 & 15.4 & 23.1 & 15.4 & 7.7 & 15.4 \\
Gemini-3.1-Pro \citep{gemini31} & 30.8 & 15.4 & 23.1 & 23.1 & 38.4 & 23.1 & 23.1 & 28.2 \\
Claude-Sonnet-4.6 \citep{claude46} & 46.2 & 46.2 & 38.5 & 43.6 & 46.2 & 53.8 & 46.2 & 48.7 \\
\midrule
\multicolumn{9}{l}{\textit{GUI-specialized Model + History-Retention Protocol (HRP)}} \\
\midrule
CogAgent-9B \citep{hong2024cogagent} & 0.0 & 0.0 & 0.0 & 0.0 & 7.7 & 0.0 & 0.0 & 2.6 \\
UI-Tars-1.5-7B \citep{qin2025ui} & 7.7 & 7.7 & 7.7 & 7.7 & 7.7 & 15.4 & 15.4 & 12.8\\
UI-Venus-1.5-8B \citep{team2026ui} & 7.7 & 0.0 & 0.0 & 2.6 & 15.4 & 7.7 & 7.7 & 10.3 \\
MAI-UI-8B \citep{zhou2025mai} & 15.4 & 7.7 & 7.7 & 10.3 & 15.4 & 15.4 & 7.7 & 12.8 \\
GUI-Owl-1.5-8B-Think \citep{xu2026mobile} & 15.4 & 7.7 & 7.7 & 10.3 & 23.1 & 15.4 & 7.7 & 15.4\\
\midrule
\multicolumn{9}{l}{\textit{Ours (Training from GUI-Owl-1.5-8B-Think)}} \\
\midrule
Stamp-GUI                          & \textbf{46.2} & \textbf{30.8} & \textbf{15.4} & \textbf{30.8} & \textbf{53.8} & \textbf{38.4} & \textbf{23.1} & \textbf{38.4} \\
\bottomrule
\end{tabular}
\caption{Main results on \textbf{AndroidWorld-M}. We report task success accuracy under pass@1 and pass@3.}
\label{tab:aw-m-main}
\end{table*}

\begin{table*}[t]
\centering
\renewcommand{\arraystretch}{0.9}
\small
\setlength{\tabcolsep}{9pt}
\begin{tabular}{l|ccc>{\columncolor{gray!15}}c|ccc>{\columncolor{gray!15}}c}
\toprule
\multirow{2}{*}{\textbf{Model}} & \multicolumn{4}{c}{\textbf{\textit{pass@1}}} & \multicolumn{4}{c}{\textbf{\textit{pass@3}}} \\
\cmidrule(lr){2-5} \cmidrule(lr){6-9}
 & L1 & L2 & L3 & Avg. & L1 & L2 & L3 & Avg. \\
\midrule
\multicolumn{9}{l}{\textit{General-purpose Multimodal Model + Active Memory Instruction (AMI)}} \\
\midrule
Qwen3-VL-8B-Instruct \citep{bai2025qwen3} & 11.1 & 0.0 & 7.4 & 6.2 & 25.9 & 7.4 & 18.5 & 17.3 \\
Qwen3-VL-8B-Think \citep{bai2025qwen3} & 18.5 & 7.4 & 3.7 & 9.9 & 40.7 & 14.8 & 14.8 &23.4 \\
GPT-5.4 \citep{gpt54} & 37.0 & 30.8 & 33.3 & 33.7 & 70.4 & 59.3 & 74.1 & 67.9 \\
Gemini-3.1-Pro \citep{gemini31} & 48.1 & 37.0 & 22.2 & 35.8 & 74.1 & 51.9 & 44.4 & 56.8 \\
Claude-Sonnet-4.6 \citep{claude46} & 66.7 & 70.4 & 51.9 & 63.0 & 77.8 & 77.8 & 70.4 & 75.3 \\
\midrule
\multicolumn{9}{l}{\textit{GUI-specialized Model + History-Retention Protocol (HRP)}} \\
\midrule
CogAgent-9B \citep{hong2024cogagent} & 0.0 & 0.0 & 0.0 & 0.0 & 0.0 & 0.0 & 0.0 & 0.0 \\
UI-Tars-1.5-7B \citep{qin2025ui} & 7.4 & 7.4 & 0.0 & 4.9 & 14.8 & 14.8 & 7.4 & 12.3 \\
UI-Venus-1.5-8B \citep{team2026ui} & 3.7 & 7.4 & 14.8 & 8.6 & 18.5 & 11.1 & 25.9 & 18.5 \\
MAI-UI-8B \citep{zhou2025mai} & 22.2 & 14.8 & 18.5 & 18.5 & 29.6 & 22.2 & 22.2 & 24.7\\
GUI-Owl-1.5-8B-Think \citep{xu2026mobile} & 22.2 & 18.5 & 18.5 & 19.7 & 25.9 & 22.2 & 25.9 & 24.7 \\
\midrule
\multicolumn{9}{l}{\textit{Ours (Training from GUI-Owl-1.5-8B-Think)}} \\
\midrule
Stamp-GUI  & \textbf{25.9} & \textbf{22.2} & \textbf{22.2} & \textbf{23.4} & \textbf{37.0} & \textbf{33.3} & \textbf{40.7} & \textbf{37.0} \\
\bottomrule
\end{tabular}
\caption{Main results on \textbf{MemGUI-Bench}. We report task success accuracy under pass@1 and pass@3.}
\label{tab:memgui-main}
\end{table*}

\begin{table*}[t]
\centering
\renewcommand{\arraystretch}{0.9}
\small
\setlength{\tabcolsep}{12pt}
\begin{tabular}{l|cc|cc}
\toprule
\multirow{2}{*}{\textbf{Model}} & \multicolumn{2}{c}{\textbf{\textit{pass@1}}} & \multicolumn{2}{c}{\textbf{\textit{pass@3}}} \\
\cmidrule(lr){2-3} \cmidrule(lr){4-5}
 & T-Acc & M-Acc & T-Acc & M-Acc \\
\midrule
\multicolumn{5}{l}{\textit{General-purpose Multimodal Model + Active Memory Instruction (AMI)}} \\
\midrule
Qwen3-VL-8B-Instruct \citep{bai2025qwen3} & 16.5 & 46.6 & 19.7 & 50.4\\
Qwen3-VL-8B-Think \citep{bai2025qwen3} & 36.2 & 60.6 & 41.7 & 64.6\\
GPT-5.4 \citep{gpt54} & 52.0 & 85.4 & 56.7 & 89.8 \\
Gemini-3.1-Pro \citep{gemini31} & 49.6 & 79.6 & 55.2 & 87.4 \\
Claude-Sonnet-4.6 \citep{claude46} & 59.1 & 95.2 & 67.7 & 98.5 \\
\midrule
\multicolumn{5}{l}{\textit{GUI-specialized Model + History-Retention Protocol (HRP)}} \\
\midrule
CogAgent-9B \citep{hong2024cogagent} & 19.7 & 24.4 & 22.1 & 26.8\\
UI-Tars-1.5-7B \citep{qin2025ui} & 20.5 & 33.9 & 26.8 & 37.4\\
UI-Venus-1.5-8B \citep{team2026ui} & 18.9 & 21.2 & 22.0 & 24.4\\
MAI-UI-8B \citep{zhou2025mai} & 40.2 & 51.2 & 45.7 & 56.7\\
GUI-Owl-1.5-8B-Think \citep{xu2026mobile} & 44.9 & 55.9 & 52.0 & 59.8\\
\midrule
\multicolumn{5}{l}{\textit{Ours (Training from GUI-Owl-1.5-8B-Think)}} \\
\midrule
Stamp-GUI & \textbf{75.3} & \textbf{77.9} & \textbf{82.6} & \textbf{84.2} \\
\bottomrule
\end{tabular}
\caption{Main results on \textbf{Memory-World}. We report memory accuracy (M-Acc) and task accuracy (T-Acc) under pass@1 and pass@3.}
\label{tab:mw-main}
\end{table*}

\begin{table*}[t]
\centering
\renewcommand{\arraystretch}{0.9}
\small
\setlength{\tabcolsep}{6.5pt}
\begin{tabular}{l|cc|cc|cccc}
\toprule
\multirow{2}{*}{\textbf{Method}} 
& \multicolumn{2}{c}{\textbf{AndroidWorld-M}} 
& \multicolumn{2}{c}{\textbf{MemGUI-Bench}} 
& \multicolumn{2}{c}{\textbf{Memory-World} \textbf{\textit{(pass@1)}}} 
& \multicolumn{2}{c}{\textbf{Memory-World} \textbf{\textit{(pass@3)}}} \\
\cmidrule(lr){2-3} \cmidrule(lr){4-5} \cmidrule(lr){6-7} \cmidrule(lr){8-9}
& \textbf{\textit{pass@1}} & \textbf{\textit{pass@3}} & \textbf{\textit{pass@1}} & \textbf{\textit{pass@3}} & T-Acc & M-Acc & T-Acc & M-Acc \\
\midrule
Baseline       & 10.3  & 15.4 & 19.7 & 24.7 & 44.9 & 55.9 & 52.0 & 59.8 \\
+ Virtual Env  & 20.8 & 31.1 & 22.2 & 30.8 & 57.5 & 68.3 & 63.1 & 71.9 \\
+ Step Balance & 28.2 & \textbf{38.4} & \textbf{23.4} & 32.1 & 69.4 & 74.8 & 73.2 & 79.4 \\
+ A/M Critic   & 25.7 & 35.9 & 21.0 & \textbf{39.5} & 74.3 & 77.4 & 81.0 & 83.6 \\
+ Online RL    & \textbf{30.8} & \textbf{38.4} & \textbf{23.4} & 37.0 & \textbf{75.3} & \textbf{77.9} & \textbf{82.6} & \textbf{84.2} \\
\bottomrule
\end{tabular}
\caption{Ablation results. We incrementally add each component in STAMP on training baseline. For AndroidWorld-M and MemGUI-Bench, we report the average task accuracy over L1--L3 under pass@1 and pass@3.}
\label{tab:ablation}
\end{table*}

\subsection{Baselines and Implementation Details}
We compare Stamp-GUI against strong GUI-specialized models and general-purpose multimodal models, including CogAgent-9B \citep{hong2024cogagent}, UI-Tars-1.5-7B \citep{qin2025ui}, UI-Venus-1.5-8B \citep{team2026ui}, MAI-UI-8B \citep{zhou2025mai}, GUI-Owl-1.5-8B-Think \citep{xu2026mobile}, Qwen3-VL-8B-Instruct \citep{bai2025qwen3}, Qwen3-VL-8B-Think \citep{bai2025qwen3}, GPT-5.4 \citep{gpt54}, Gemini-3.1-Pro \citep{gemini31}, and Claude-Sonnet-4.6 \citep{claude46}. For fair comparison, we use a best-effort memory elicitation setup tailored to each model family. 
For general-purpose multimodal models, we append an explicit instruction asking the model to output information that may need to be remembered for later use. 
For GUI-specialized models, we found that most are tightly coupled to their training-time prompting formats and generally do not reliably follow additional memory-specific instructions. 
Accordingly, rather than forcing extra instructions that are often ignored, we evaluate them with their native prompting interface and adopt a simple History-Retention Protocol (HRP): we preserve the full dialogue history throughout the rollout, and the task format encourages key task-relevant information to be expressed in the model's textual responses whenever possible. 
This setup is intended to provide a best-effort, instruction-free way for GUI-specialized models to retain recoverable information from prior steps.

For M-Acc, because most GUI-specialized models do not expose a dedicated memory field, we score memory by matching against the retained interaction history and model outputs, giving each baseline the maximum recoverable credit from its produced trajectory. 
We include frontier general-purpose multimodal models as strong reference baselines, but our primary comparison is with GUI-specialized models, since STAMP targets native memory behavior within the mobile interaction loop.

For implementation, we synthesize 4,194 trajectories via our dynamic testing pipeline for supervised fine-tuning, and configure 356 virtual environments for online reinforcement learning. Gemini-2.5-Pro \citep{comanici2025gemini} serves as the environment generator, Claude-Sonnet-4.5 \citep{claude45} acts as the step-level critic, and Qwen-3.5 \citep{qwen35} is employed as the judge for the RL memory reward. Stamp-GUI is initialized from GUI-Owl-1.5-8B-Think.

\subsection{Main Results}
As compiled in Table~\ref{tab:aw-m-main}, \ref{tab:memgui-main}, and \ref{tab:mw-main}, Stamp-GUI consistently outperforms prior GUI-specialized models across all three benchmarks and achieves the best overall performance on Memory-World. On AndroidWorld-M and MemGUI-Bench, several much larger frontier general-purpose models remain stronger under explicit memory prompting, but Stamp-GUI sets a new state of the art within the GUI-specialized model family. We note that several frontier closed-source multimodal models outperform GUI-specialized models under active-memory prompting, which is expected given their substantially larger scale. Our focus, however, is on improving GUI-specialized models through explicit memory training rather than competing purely via model size. A deeper analysis of the results reveals three key insights:

\paragraph{Consistent Dominance Across Evaluation Levels.} 
The empirical results across the three-level difficulty protocol underscore Stamp-GUI's comprehensive superiority. In the oracle-assisted L1 and L2 tiers, our model sets a remarkably high performance ceiling, demonstrating an advanced capacity to leverage explicit context. When evaluated under the unassisted L3 protocol, which demands absolute autonomy, standard agents collapse due to the permanent loss of historical visual cues. Conversely, Stamp-GUI demonstrates robust resistance to this difficulty scaling. This confirms that the integration of an explicit memory field provides a stable information foundation.

\paragraph{Memory Fidelity as a Foundation for Task Success.} 
On the Memory-World benchmark, Stamp-GUI achieves a commanding lead in M-Acc alongside top-tier T-Acc scores. This strong correlation demonstrates that our model does not solve tasks through random exploration or shortcuts; rather, it systematically records the correct transient parameters into its explicit memory field, providing a stable information foundation for executing long-horizon downstream actions.

\paragraph{Multi-Attempt Scalability.} 
Across all evaluated environments, the improvement of Stamp-GUI from pass@1 to pass@3 is more pronounced than that of competing baselines. This suggests that explicit memory improves robustness under repeated attempts and makes task completion less brittle in long-horizon settings. While pass@k alone does not establish a specific correction mechanism, the stronger scaling indicates that the model can benefit more consistently from multiple rollouts.

\begin{figure*}[t]
    \centering
    \includegraphics[width=1\linewidth]{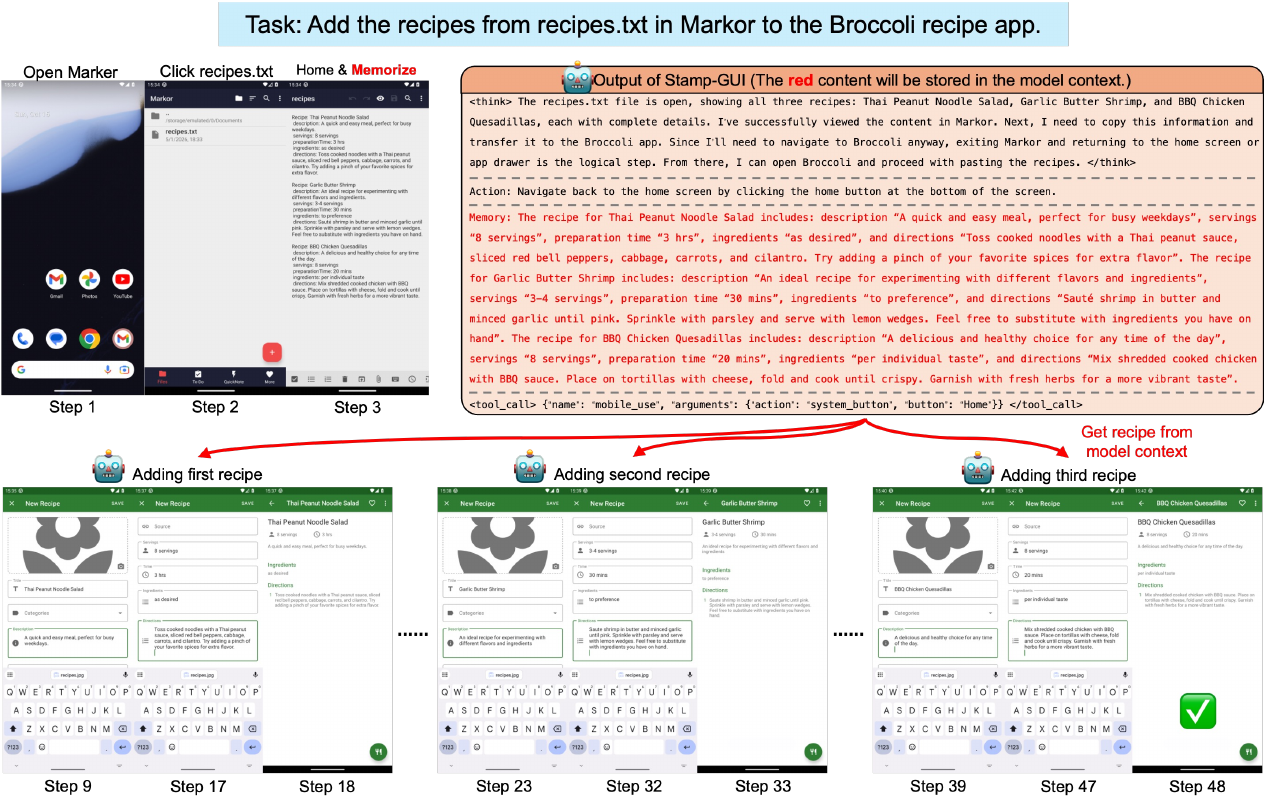}
    \caption{Example of Stamp-GUI predict memory content when appropriate and using memory when needed on an L3-level task from AndroidWorld-M.}
    \label{fig:case}
    \vspace{-3mm}
\end{figure*}

\subsection{Ablation Study}
We conduct an incremental ablation study to verify the individual contributions of the core STAMP components. As shown in Table~\ref{tab:ablation}, our ablation results indicate that incorporating automated virtual environment trajectories yields the largest foundational improvement, confirming the effectiveness of our synthesized memory-intensive training tasks. Introducing the step balancing mechanism prevents the model from ignoring the memory field, leading to noticeable boosts on real-world benchmarks. Furthermore, the step-level critic introduces a non-uniform trade-off: although it slightly lowers some aggregate scores on AndroidWorld-M and MemGUI-Bench, it consistently improves the memory-focused metrics, indicating better supervision fidelity. Online RL then compensates for these regressions and yields the best overall balance.

\subsection{Detailed Empirical Analysis}

\paragraph{Qualitative Case Study.}
To illustrate the operational mechanism of Stamp-GUI, we analyze a complex cross-app task from the Markor text editor to the Broccoli recipe. Standard end-to-end agents typically fail this task because the detailed recipe information is completely purged from the visual context upon switching apps. As shown in Figure~\ref{fig:case}, Stamp-GUI resolves this bottleneck by proactively extracting and compiling the recipe details into its explicit \texttt{Memory} output field while interacting with Markor. Upon navigating to the Broccoli app, this memory remains securely within the model's history context, allowing the agent to accurately retrieve the transient data.

\begin{figure}[!ht]
    \centering
    \includegraphics[width=1\linewidth]{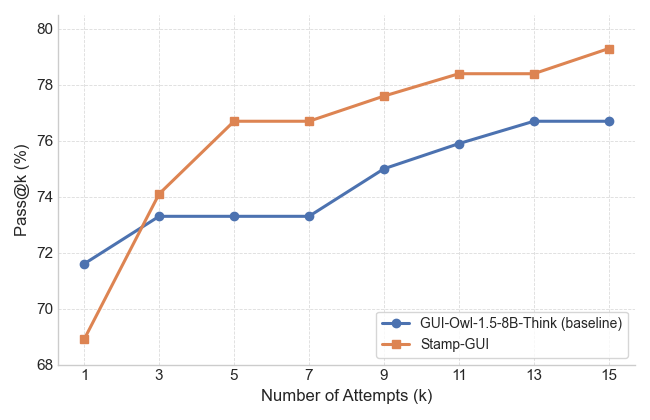}
    \caption{General capability evaluation on the full 116 AndroidWorld evaluation tasks.}
    \label{fig:androidworld-general}
\end{figure}

\paragraph{General Mobile Capability.}
To ensure that optimizing for memory does not degrade the agent's general navigation capabilities, we evaluate Stamp-GUI on the full AndroidWorld benchmark. We plot the pass@$k$ scaling curve from $k=1$ to $15$ in Figure~\ref{fig:androidworld-general}. While Stamp-GUI experiences a minor degradation at pass@1 compared to its base model due to the formatting constraints of explicit memory generation, its scaling trajectory is significantly steeper. It surpasses the base model at pass@3 and maintains a steady lead up to pass@15, achieving a final success rate of 79.3\%. This confirms that explicit memory training ultimately enhances long-horizon robustness without harming general execution skills.

\paragraph{Effect of Step Balancing.}
We examine the impact of data formulation strategies during supervised fine-tuning. As shown in Table~\ref{tab:step-balance}, our hyperparameter analysis on the step balancing mechanism demonstrates that shifting from the raw, highly skewed vanilla distribution to an upsampled configuration significantly enhances policy performance. By explicitly elevating the proportion of memory-active steps within the training sequence, the model receives denser gradient signals associated with internal state tracking. This data-level adjustment effectively guides the network to robustly learn when and what to record in memory.

\begin{table}[!ht]
\centering
\renewcommand{\arraystretch}{0.9}
\small
\setlength{\tabcolsep}{4.5pt}
\begin{tabular}{l|c|c|cc}
\toprule
\multirow{2}{*}{\textbf{Method}} & \textbf{AW-M} & \textbf{MemGUI} & \multicolumn{2}{c}{\textbf{Memory-World}} \\
\cmidrule(lr){2-2} \cmidrule(lr){3-3} \cmidrule(lr){4-5}
 & Avg. & Avg. & T-Acc & M-Acc \\
\midrule
Vanilla   & 20.8/31.1 & 22.2/30.8  & 57.5/63.1 & 68.3/71.9 \\
Up to 1:1 & 25.7/\textbf{41.0} & 22.2/34.6 & 71.7/78.4 & 73.3/81.1 \\
Up to 3:1 & \textbf{30.8}/38.4 & \textbf{23.4}/\textbf{37.0} & \textbf{75.3}/\textbf{82.6} & \textbf{77.9}/\textbf{84.2} \\
\bottomrule
\end{tabular}
\caption{Effect of step balance. We report the average accuracy over L1--L3 under pass@1/pass@3.}
\label{tab:step-balance}
\vspace{-3mm}
\end{table}

\paragraph{Dynamic testing quality.}
We assessed the robustness of the dynamic testing phase across different planner models. As detailed in Table~\ref{tab:dynamic-testing}, our pipeline maintains high success rates across various models, requiring an efficient number of interaction steps. Crucially, guided by the generated solution documents, the proportion of logically correct steps approaches 90\% regardless of the model used. This confirms that STAMP's framework is not overly reliant on any single proprietary model.

\begin{table}[!ht]
\centering
\renewcommand{\arraystretch}{0.9}
\small
\setlength{\tabcolsep}{6.5pt}
\begin{tabular}{l|ccc}
\toprule
\textbf{Model} & \textbf{SR} & \textbf{Avg. Step} & \textbf{Correct Step} \\
\midrule
GPT-5.4    & 69.8 & 15.4 & 83.6 \\
Gemini-3.1-Pro   & 74.5 & 13.2 & 87.2 \\
Claude-Sonnet-4.6 & 77.4 & 12.7 & 89.5 \\
\bottomrule
\end{tabular}
\caption{Dynamic testing results. SR denotes success rate, Avg. Step denotes the average steps, and Correct Step denotes the proportion of correct steps.}
\label{tab:dynamic-testing}
\vspace{-3mm}
\end{table}

\section{Conclusion}

This paper addresses a fundamental bottleneck in mobile GUI agents: the loss of transient information due to the necessity of discarding historical screenshots. We present STAMP, a scalable framework that shifts training from static, action-centric data to programmatically generated virtual environments. By synthesizing tasks with deterministic memory variables, STAMP enables agents to learn explicit memory prediction through both supervised fine-tuning and online reinforcement learning. Our experiments demonstrate that Stamp-GUI significantly improves performance in memory scenarios, proving that explicit memory is essential for building resilient, scalable mobile agents.

\section{Limitations}

Our work has several limitations. Although STAMP uses controllable and scalable virtual environments to generate executable and automatically verifiable memory supervision, these environments still cannot fully capture the diversity, noise, and hidden dynamics of real mobile applications. In addition, the current pipeline relies on multiple strong models, which improves automation quality but also increases computational and engineering cost. Finally, our current formulation does not yet cover broader forms of memory such as long-term user preferences, cross-task memory, or more implicit state tracking. These limitations suggest promising future directions in improving virtual-to-real transfer, reducing pipeline cost, and broadening the scope of memory modeling and evaluation.

\bibliography{custom}

\clearpage

\appendix

\section{Appendix}
\label{sec:appendix}

\subsection{Output Format}
\label{sec:format}
Stamp-GUI extends the standard output format of GUI-Owl-1.5 series by adding an explicit memory field. The original format is:
\begin{verbatim}
<think> think content </think>
Action: action description
<tool_call> action content </tool_call>
\end{verbatim}
Our format is:
\begin{verbatim}
<think> think content </think>
Action: action description
Memory: memory content
<tool_call> action content </tool_call>
\end{verbatim}

\subsection{Memory Content Storage}
\label{app:memory_content_storage}

This section describes how we store interaction history when the model is required to predict an explicit \texttt{Memory} field at every step. Our goal is to preserve information that is most useful for downstream execution while keeping the prompt length bounded. In particular, we treat \texttt{Action} and \texttt{Memory} as the two most important textual carriers of long-horizon state, and we store them with higher fidelity than the free-form reasoning content inside \texttt{<think>}.

\paragraph{Step output format.}
At each interaction step $t$, the model produces a structured response as shown in~\ref{sec:format}. Let the generated output at step $t$ be denoted by
\begin{equation}
y_t = \big(c_t, a_t, m_t, u_t\big),
\end{equation}
where $c_t$ is the content inside \texttt{}, $a_t$ is the textual action description, $m_t$ is the memory content, and $u_t$ is the tool-call payload.

\subsubsection{Two-Tier History Retention}

Suppose the current decision step is $t$. We divide previous interaction history into two segments:

\begin{itemize}
    \item \textbf{Recent window:} the most recent $K=5$ rounds, i.e.,
    \[
    \mathcal{R}_t = \{t-5, t-4, \dots, t-1\},
    \]
    where valid indices are clipped to be positive.
    These rounds are preserved in their original full format, including the complete user/assistant interaction structure.

    \item \textbf{Early window:} all rounds earlier than the recent window, i.e.,
    \[
    \mathcal{H}_t = \{1,2,\dots,t-6\}.
    \]
    For these older rounds, we do not keep the full raw dialogue. Instead, we extract only the \texttt{Action} and \texttt{Memory} fields and re-organize them into a compact stepwise history block.
\end{itemize}

Therefore, the model input at step $t$ is composed of three parts:
\begin{equation}
X_t = \big(q,\; \texttt{History}_{1:t-6},\; \texttt{RecentTurns}_{t-5:t-1},\; s_t\big),
\end{equation}
where $q$ is the user instruction, $\texttt{History}_{1:t-6}$ is the compressed action-memory history, $\texttt{RecentTurns}_{t-5:t-1}$ contains the last five full rounds, and $s_t$ denotes the current observation and current user-side input for the next decision.

\subsubsection{History Construction from Old Steps}

For each old step $i \in \mathcal{H}_t$, we convert the original output into a compact textual record:
\begin{equation}
h_i = \texttt{Step }i\texttt{: } a_i \;\Vert\; \texttt{\textbackslash nMemory: } m_i,
\end{equation}
where $\Vert$ denotes string concatenation.

The complete compressed history is then
\begin{equation}
\texttt{History}_{1:t-6}
=
h_1 \;\Vert\; \texttt{\textbackslash n} \;\Vert\; h_2 \;\Vert\; \cdots \;\Vert\; \texttt{\textbackslash n} \;\Vert\; h_{t-6}.
\end{equation}

In other words, older history is serialized as:
\begin{verbatim}
Step 1: action description
Memory: memory content
Step 2: action description
Memory: memory content
...
Step t-6: action description
Memory: memory content
\end{verbatim}

\subsubsection{Storage Behavior for Empty Memory}

Not every step requires memorization. For steps with no memory prediction, we set $m_i=\varnothing$. In implementation, this can be represented in either of the following equivalent ways:
\begin{verbatim}
    Step i: action description
    Memory: none
\end{verbatim}

\subsection{Benchmark Details}
\label{app:benchmark_details}

In this section, we provide the task-level details for the benchmarks used in our experiments. Since AndroidWorld-M and MemGUI-Bench are built upon publicly available benchmarks and we only introduce controlled memory-oriented modifications, we first document the specific task configurations and edits applied to these two benchmarks. We defer the details of our newly introduced Memory-World benchmark to the next subsection.

\subsubsection{AndroidWorld-M}
\label{app:androidworld_m_details}

AndroidWorld-M contains three difficulty levels, denoted as L1, L2, and L3. Each difficulty level includes 13 tasks. Among them, 8 are native tasks from AndroidWorld and 5 are extended variants constructed by modifying native tasks into single-instance memory tasks. For each task, we report: (i) the task identifier, (ii) whether it is native or extended, (iii) the instruction used for L1/L2, (iv) the instruction used for L3, and (v) the specific simplification applied in L1 to reduce task difficulty.

\subsubsection{MemGUI-Bench}
\label{app:memgui_bench_details}

For MemGUI-Bench, we follow the same three-level evaluation protocol (L1/L2/L3) used in AndroidWorld-M. To clearly present the benchmark configuration, we list the task description for each task identifier under all three difficulty levels. Specifically, L1 contains the easiest task formulation, L2 shifts some tasks from full action execution to direct-answer variants while preserving the memory requirement, and L3 uses the most natural and minimally guided formulations.

\subsection{Prompt Templates}
\label{app:prompt_templates}

This section presents the prompts used in our agent. We use a fixed system prompt to define the tool interface and output format, and a user prompt to provide the current task instruction together with the retained interaction history. As described in Appendix~\ref{app:memory_content_storage}, each user turn also contains the current UI screenshot. During inference, only the screenshots from the most recent five user turns are retained; earlier turns are compressed into textual history and their images are removed.

\subsubsection{System Prompt}

Table~\ref{tab:system_prompt} shows the full system prompt used in our experiments.

\subsubsection{User Prompt}

Table~\ref{tab:user_prompt} shows the user prompt template. The placeholders \texttt{\{goal\}} and \texttt{\{history\_string\}} are replaced by the task instruction and the retained history string, respectively.

\begin{table}[!ht]
\centering
\small
\renewcommand{\arraystretch}{1}
\begin{tabular}{p{0.45\textwidth}}
\toprule
\textbf{User Prompt} \\
\midrule
\ttfamily
Please generate the next move according to the UI screenshot, instruction and previous actions. \\
\\
Instruction: \{goal\} \\
\\
Previous actions: \\
\{history\_string\} \\
\bottomrule
\end{tabular}
\caption{User prompt template used by Stamp-GUI.}
\label{tab:user_prompt}
\end{table}

\subsubsection{Prompt Assembly with Visual History}

At step $t$, the input to the model consists of:
\begin{equation}
\begin{split}
\mathcal{P}_t = \big(&
\texttt{SystemPrompt},\;
\texttt{UserTurns}_{t-4:t},\\
&\texttt{CompressedHistory}_{1:t-5}
\big),
\end{split}
\end{equation}

where $\texttt{SystemPrompt}$ is the fixed prompt in Table~\ref{tab:system_prompt}, $\texttt{UserTurns}_{t-4:t}$ denotes the most recent five user turns, each containing its corresponding screenshot, $\texttt{CompressedHistory}_{1:t-5}$ denotes the older history converted into the action-memory format described in Appendix~\ref{app:memory_content_storage}.

More concretely, each user turn includes both text and one screenshot image. To control the context length, we only retain screenshots from the most recent five user turns. This design ensures that the model retains recent visual grounding while carrying forward older task-relevant information only through the explicit \texttt{Action} and \texttt{Memory} fields.

\subsection{Reinforcement Learning Details}
\label{app:rl_details}

This section summarizes the reinforcement learning setup used for Stamp-GUI. We focus on the main training configuration, rollout pipeline, and advantage computation strategy, while omitting low-level implementation details.

\subsubsection{Training Framework}

We perform online reinforcement learning with a PPO-style policy optimization framework built on top of a distributed rollout system. The policy is initialized from the supervised fine-tuned checkpoint and further optimized through interaction with executable virtual environments. In our implementation, environment interaction, rollout generation, and policy updates are decoupled through a remote asynchronous pipeline, allowing environment execution to proceed continuously while model updates are performed in parallel.

\subsubsection{Distributed Rollout Configuration}

We use a remote rollout backend with asynchronous task execution. The rollout worker generates multiple candidate trajectories per task and returns them to the trainer through a bounded buffer. The implementation enforces strict on-policy sampling and supports task rollback when environment execution becomes invalid. To improve hardware utilization, the rollout stage uses buffered scheduling and asynchronous task upload.

\subsubsection{Policy Optimization Hyperparameters}

We optimize the actor with PPO-style clipped policy updates. The training configuration is intentionally conservative because the model already has a strong supervised initialization and the RL stage mainly serves as behavior refinement. We additionally apply KL regularization against the reference policy to prevent reward hacking and to maintain stable action formatting. The main optimization hyperparameters are listed in Table~\ref{tab:rl_optim_config}.

\begin{table}[!ht]
\centering
\small
\renewcommand{\arraystretch}{1}
\setlength{\tabcolsep}{15pt}
\begin{tabular}{lc}
\toprule
\textbf{Item} & \textbf{Value} \\
\midrule
Optimizer learning rate & $2\times 10^{-7}$ \\
Train batch size & 32 \\
PPO mini-batch size & 32 \\
Micro-batch size per GPU & 1 \\
Policy loss mode & PPO \\
Clip ratio (low / high) & 0.05 / 0.05 \\
KL loss & Enabled \\
KL coefficient & 0.05 \\
KL type & Low-variance KL \\
Entropy coefficient & $5\times 10^{-4}$ \\
Gradient checkpointing & Enabled \\
Critic warmup & 0 \\
\bottomrule
\end{tabular}
\caption{Main optimization hyperparameters used in online RL.}
\label{tab:rl_optim_config}
\end{table}

\subsubsection{Step-Level GRPO Advantage}

Our implementation uses a step-aware grouped relative policy optimization estimator, denoted as \texttt{step\_grpo}. The central motivation is that GUI-agent trajectories are multi-step and strongly correlated within the same task instance. Instead of treating each sampled sequence independently, we normalize outcomes within structured task groups.

Let $s_i$ denote the scalar score of sample $i$, obtained by summing its token-level rewards over the response:
\begin{equation}
s_i = \sum_{k} r_{i,k}.
\end{equation}
Each sample is associated with a task ID, a trajectory ID, and a step ID:
\[
(\texttt{task\_id}_i,\; \texttt{traj\_id}_i,\; \texttt{step\_id}_i).
\]

We support several grouping strategies in the implementation, but the core idea is to compute a normalized outcome score within a task-conditioned group and then broadcast that signal back to the steps of the trajectory.

\paragraph{Last-step grouping.}
The default variant first identifies the final step of each sampled trajectory, uses only these terminal outcomes to compute task-level statistics, and then broadcasts the resulting normalized score back to all steps from the same trajectory. Formally, for each trajectory $(u,v)$ indexed by task $u$ and trajectory $v$, let $i^\star(u,v)$ be its final step. We compute
\begin{equation}
\hat{s}_{u,v}
=
\frac{s_{i^\star(u,v)} - \mu_u}{\sigma_u + \epsilon},
\end{equation}
where $\mu_u$ and $\sigma_u$ are the mean and standard deviation of final-step scores across trajectories belonging to task $u$. This normalized terminal outcome is then assigned to all steps in the same trajectory:
\begin{equation}
A_i = \hat{s}_{u,v}, \qquad \forall i \in (u,v).
\end{equation}

This strategy is especially suitable when the final environment outcome is the most reliable indicator of whether the step sequence was useful.

\paragraph{Optional temporal decay.}
We also support a simple decay factor when propagating final-step credit backward:
\begin{equation}
A_i = \hat{s}_{u,v} \cdot \beta^{(t_{\max}-t_i)},
\end{equation}
where $t_i$ is the step index, $t_{\max}$ is the final step of the trajectory, and $\beta \in (0,1]$ is a decay coefficient. In our reported setting, the decay coefficient is fixed to $1.0$, i.e., no additional temporal attenuation is applied.

\paragraph{Each-step grouping.}
A second variant computes normalization separately for each \texttt{(task\_id, step\_id)} group:
\begin{equation}
A_i
=
\frac{s_i - \mu_{u,t}}{\sigma_{u,t} + \epsilon},
\end{equation}
where $\mu_{u,t}$ and $\sigma_{u,t}$ are computed over all samples corresponding to the same task $u$ and the same step index $t$. This produces a more local comparison signal, but in our setup the trajectory-level final-outcome grouping is the primary design.

\subsection{Trajectory Data Format}
\label{app:trajectory_data_format}

This section describes the serialized trajectory format used in STAMP for supervised training data construction and post-processing. Each trajectory contains the task instruction and a sequence of step records. The format is designed to preserve the full multimodal interaction trace, including screenshots, action annotations, reasoning traces, and explicit memory targets.

\subsubsection{Top-Level Structure}

A single trajectory instance is represented as:
\begin{verbatim}
{"goal": ..., "steps": [...]}
\end{verbatim}
Formally, we denote one trajectory instance as
\begin{equation}
\mathcal{D} = \big(q,\; \{z_t\}_{t=1}^{T}),
\end{equation}
where $q$ is the task instruction (\texttt{goal}), $z_t$ is the serialized record of step $t$, $T$ is the total number of steps.

\subsubsection{Step-Level Structure}

Each step is stored as a structured object:
\begin{equation}
z_t = \big(s_t,\; a_t,\; c_t,\; d_t,\; b_t,\; m_t\big),
\end{equation}
where $s_t$ is the screenshot reference, $a_t$ is the executable action content, $c_t$ is the long-form reasoning text, $d_t$ is the short decision summary, $b_t \in \{0,1\}$ indicates whether the step is trainable,$m_t$ is the explicit memory content.

\subsubsection{Field Definitions}

\paragraph{Action content.}
In general, we denote this field by
\begin{equation}
a_t = \big(\texttt{action}, \texttt{args}\big),
\end{equation}
where \texttt{action} specifies the operation type (e.g., click, type, terminate), and \texttt{args} contains the corresponding parameters such as coordinates or input text.

\paragraph{Thought.}
The \texttt{thought} field contains the long-form reasoning trace used during data generation. This field records the model-side reasoning that explains the current state, decomposes the task, and motivates the next action.

\paragraph{Conclusion.}
The \texttt{conclusion} field is a short textual summary of the immediate next operation. Compared with \texttt{thought}, this field is more concise and directly aligned with the final action choice.

\paragraph{Trainable flag.}
The \texttt{trainable} field indicates whether the step should be used as a supervised training target. Some steps may be marked as non-trainable if they are considered erroneous, noisy, or unsuitable for optimization. In the example provided, one correction step after an incorrect form entry is marked as \texttt{false}, meaning it can be excluded from standard supervised learning.

\paragraph{Memory.}
The \texttt{memory} field stores the explicit memory target associated with the current step. If no memorization is required, this field may be an empty string.

\subsection{Virtual Environment Generation Settings}
\label{app:virtual_env_generation_settings}

In this section, we provide representative examples of the seed configurations used in our virtual environment generation pipeline.

\subsubsection{Platform Seed}

A platform seed defines the high-level properties of the synthesized application, including its domain, visual theme, navigation structure, common entity types, and distracting elements. These attributes are used by the environment generation agent to produce a coherent mobile application with stable style and behavior.

Table~\ref{tab:platform_seed_example} shows one example platform seed used in our pipeline.

\begin{table}[!ht]
\centering
\small
\renewcommand{\arraystretch}{1}
\begin{tabular}{p{2cm}p{4.8cm}}
\toprule
\textbf{Field} & \textbf{Value} \\
\midrule
Type & \texttt{social} \\
App name & \texttt{StarWave} \\
Brand slogan & Discover every trend in everyday life \\
Style & \texttt{[Feed, Search, Profile, Messages, Me]} \\
Primary color & \texttt{\#FF5C8A} \\
Secondary color & \texttt{\#7A3CFF} \\
Background color & \texttt{\#F7F8FA} \\
Card style & Rounded cards \\
Icon style & Soft skeuomorphic \\
Common entities & \texttt{[creator, topic, hot post, followers, likes, shares]} \\
Detail entry points & \texttt{[post detail, topic detail, pinned post on creator profile]} \\
Distractions & \texttt{[Trending Picks, You May Like, Upgrade to Premium, Find Contacts, Live Stream]} \\
Text tone & Young, social, energetic \\
\bottomrule
\end{tabular}
\caption{Example of a platform seed used for virtual environment generation.}
\label{tab:platform_seed_example}
\end{table}

In this example, the generated application belongs to a social-media-like domain and is instantiated as a fictional app named \texttt{StarWave}. The environment generator uses this seed to control both visual style and semantic content. For instance, the bottom navigation bar is expected to contain tabs such as \texttt{Feed}, \texttt{Search}, \texttt{Profile}, \texttt{Messages}, and \texttt{Me}, while the page contents are populated with social entities such as creators, topics, and engagement statistics. At the same time, additional distracting modules such as \texttt{Trending Picks} or \texttt{Live Stream} can be inserted to make the environment more realistic and to increase memory interference.

\subsubsection{Task Seed}

A task seed specifies the abstract task pattern that will be instantiated inside the generated application. It controls what information the agent must extract, how many steps are required, what kind of memory burden is imposed, and what output format is expected.

Table~\ref{tab:task_seed_example} presents one example task seed.

\begin{table}[!ht]
\centering
\small
\renewcommand{\arraystretch}{1}
\begin{tabular}{p{2cm}p{4.8cm}}
\toprule
\textbf{Field} & \textbf{Value} \\
\midrule
ID & \texttt{date\_compare\_latest} \\
Description & Read dates from multiple detail pages, find the latest one, and submit the corresponding name \\
Goal & Compare date fields and output the name associated with the latest date \\
Required steps & 4 \\
Memory load & Medium \\
Output format & Output only the latest item's name \\
Answer type & \texttt{single\_label} \\
UI pattern & \texttt{[Detail A, Detail B, Detail C, Submit name]} \\
\bottomrule
\end{tabular}
\caption{Example of a task seed used for virtual task instantiation.}
\label{tab:task_seed_example}
\end{table}

This seed defines a comparison task in which the agent must visit several detail pages, read the date displayed on each page, determine which date is the latest, and finally submit the corresponding item name. Although the final answer is short, successful completion requires explicit cross-page memory because the relevant date values are distributed across multiple screens and must be compared after navigation.

\subsubsection{Seed Combination and Environment Instantiation}

In the generation pipeline, a platform seed and a task seed are combined to produce a concrete executable task instance. Let $z^{\text{plat}}$ denote the platform seed and $z^{\text{task}}$ denote the task seed. The environment generator maps them into a runnable application-task pair:
\begin{equation}
(E, q, r) = G_{\text{env}}(z^{\text{plat}}, z^{\text{task}}),
\end{equation}
where $E$ is the executable virtual environment, $q$ is the final natural-language instruction shown to the agent, and $r$ is the internal reference solution used by the dynamic testing agent.

Using the two seeds above, one possible instantiated task could be described as follows:

\begin{quote}
In the \texttt{StarWave} app, open the detail pages of three candidate creators or trending posts, compare their displayed publish dates, and submit the name corresponding to the most recent date.
\end{quote}

Under this instantiation, the task generator may place the three date fields in different entry points, such as a post detail page, a topic detail page, and a pinned post on a creator profile. The platform seed ensures that these pages are visually and semantically consistent with the \texttt{StarWave} application style, while the task seed ensures that the interaction pattern follows the desired \texttt{Detail A $\rightarrow$ Detail B $\rightarrow$ Detail C $\rightarrow$ Submit} structure.

\subsection{Agent Prompts}
\label{app:agent_prompts}

In this section, we list the main prompts used by different agents in the STAMP pipeline. These prompts correspond to the key stages of our automatic data construction and evaluation process, including dynamic testing, memory-accuracy judgment, scenario generation, task specification generation, webpage generation, and static semantic inspection. For clarity, we present each prompt in a single-column or double-column large-format table depending on its length.

\begin{table}[!ht]
\centering
\small
\renewcommand{\arraystretch}{1.2}
\begin{tabular}{p{0.945\columnwidth}}
\toprule
\textbf{Dynamic Testing Worker Prompt} \\
\midrule
\ttfamily
You are an expert in mobile phone operations. I will provide you with a mobile screenshot and an operation task, and you need to output the exact action to perform. \\
\\
Mobile screenshot: shown in the image below \\
Operation task: \{next\_planning\} \\
\\
Your action space is as follows. You must choose one of the following actions and output it in the required format: \\
- click (x, y): click the position whose coordinates are (x, y), where both x and y are relative coordinates in the range 0-1000 \\
- type (x, y, text): first click the position whose coordinates are (x, y), then input text, where text is the content to type \\
- scroll: scroll the screen downward a bit to view more content that is not currently visible below \\
\\
Your output format should follow the examples below: \\
click (500, 500) \\
type (100, 100, hello) \\
scroll \\
\\
Note: Please do not attempt to manipulate elements at the very top of the page (the browser's fixed tag bar, y<=120). All your actions should be concentrated in the webpage content area. \\
\\
Do not output anything other than the formats above. You may output only one action at a time. Note that type (x, y, text) represents two actions: first clicking (x, y), then typing text. \\

\bottomrule
\end{tabular}
\caption{Prompt used by the dynamic testing worker.}
\label{tab:prompt_dynamic_worker}
\end{table}

\begin{table*}[t]
\centering
\scalebox{0.8}{
\begin{minipage}{1.25\textwidth}
\scriptsize
\setlength{\tabcolsep}{3pt}
\renewcommand{\arraystretch}{1.15}
\begin{tabular}{p{4cm}p{1.5cm}p{6.4cm}p{5cm}p{1.5cm}}
\toprule
\textbf{Task ID} & \textbf{Type} & \textbf{L1 \& L2 Instruction} & \textbf{L3 Instruction} & \textbf{L1 Simplification Note} \\
\midrule

BrowserMultiply 
& Native 
& Open the file \texttt{task.html} in Downloads in the file manager; when prompted open it with Chrome. Then click the button that says ``click me'' 5 times, remember the 5 numbers displayed (may be same), and then make sure you see the submit page after clicking 5 times. Finally, submit the product of the 5 numbers in the form. 
& Open the file \texttt{task.html} in Downloads in the file manager; when prompted open it with Chrome. Then click the button 5 times, remember the numbers displayed, and enter their product in the form. 
& Replace the 5 numbers with fixed values 1, 2, 3, 4, 5. \\
\midrule

ExpenseAddSingleFromGallery 
& Ext. 
& Go to Simple Gallery Pro and search for \texttt{expenses.jpg} in DCIM to remember the expenses on \texttt{expenses.jpg} (first go to the DCIM folder, then use the search bar to search for \texttt{expenses.jpg}, finally make sure you are looking at \texttt{expenses.jpg} and remember the three expenses). Then add the remembered three expenses to \{\_APP\_NAME\} (make sure each expense has correct Name, Amount, category, and Note). 
& Add the expenses from \texttt{expenses.jpg} in Simple Gallery Pro to \{\_APP\_NAME\}. 
& Reduce Gallery noise images from 10 to 3. \\
\midrule

ExpenseAddMultipleFromGallery 
& Native 
& Go to Simple Gallery Pro and search for \texttt{expenses.jpg} in DCIM to remember the expenses on \texttt{expenses.jpg} (first go to the DCIM folder, then use the search bar to search for \texttt{expenses.jpg}, finally make sure you are looking at \texttt{expenses.jpg} and remember the three expenses). Then add the remembered three expenses to \{\_APP\_NAME\} (make sure each expense has correct Name, Amount, category, and Note). 
& Add the expenses from \texttt{expenses.jpg} in Simple Gallery Pro to \{\_APP\_NAME\}. 
& Reduce target expenses from 3 to 2; Gallery noise from 10 to 3. \\
\midrule

ExpenseAddSingleFromMarkor 
& Ext. 
& Go to Markor and find \texttt{my\_expenses.txt} to remember the one reimbursable transaction (whose note includes ``Reimbursable.''; if you do not find the reimbursable transaction, please scroll the page). Then add the remembered reimbursable transaction to \{\_APP\_NAME\} (make sure each expense has correct Name, Amount, category, and Note). 
& Go through the transactions in \texttt{my\_expenses.txt} in Markor. Log the reimbursable transactions in the \{\_APP\_NAME\}. 
& Reduce Markor noise entries from 100 to 1. \\
\midrule

ExpenseAddMultipleFromMarkor 
& Native 
& Go to Markor and find \texttt{my\_expenses.txt} to remember the one reimbursable transaction (whose note includes ``Reimbursable.''; if you do not find the reimbursable transaction, please scroll the page). Then add the remembered reimbursable transaction to \{\_APP\_NAME\} (make sure each expense has correct Name, Amount, category, and Note). 
& Go through the transactions in \texttt{my\_expenses.txt} in Markor. Log the reimbursable transactions in the \{\_APP\_NAME\}. 
& Reduce target expenses from 3 to 2; Markor noise from 100 to 1. \\
\midrule

MarkorTranscribeReceipt 
& Native 
& Go to Simple Gallery app to view the \texttt{receipt.png}, and remember the receipt on it. Then go to Markor app to create a new file called \texttt{receipt.md} and enter transactions in csv format including the header ``Date, Item, Amount''. 
& Create a file in Markor, called \texttt{receipt.md} with the transactions from the \texttt{receipt.png}. Use Simple Gallery to view the receipt. Please enter transactions in csv format including the header ``Date, Item, Amount''. 
& Increase max step limit from 18 to 25. \\
\midrule

MarkorTranscribeVideo 
& Native 
& Transcribe the contents of video \texttt{\{video\_name\}} by watching it in VLC player (located in Download) and writing the sequence of strings shown on each frame to the text file \texttt{\{file\_name\}} in Markor as a comma separated list. 
& Transcribe the contents of video \texttt{\{video\_name\}} by watching it in VLC player (located in Download) and writing the sequence of strings shown on each frame to the text file \texttt{\{file\_name\}} in Markor as a comma separated list. For example, if the first frame shows the text ``edna'' and the second frame shows the text ``pineapple'', then the text file should contain only the following text: ``edna, pineapple''. 
& Increase max step limit from 20 to 40. \\
\midrule

RecipeAddSingleRecipesFromImage 
& Ext. 
& Go to Simple Gallery Pro and open \texttt{recipes.jpg} in DCIM to remember all the information of the three recipes (first go to the DCIM folder, then just click the \texttt{recipes.jpg} on the top left once to open it, finally make sure you are looking at \texttt{recipes.jpg} and remember the three recipes). Then add the remembered three recipes to Broccoli recipe app. Please make sure the name, description, servings, preparation time, ingredients, and directions are entered correctly. 
& Add the recipes from \texttt{recipes.jpg} in Simple Gallery Pro to the Broccoli recipe app. 
& Reduce Gallery noise images from 10 to 1. \\
\midrule

RecipeAddMultipleRecipesFromImage 
& Native 
& Go to Simple Gallery Pro and open \texttt{recipes.jpg} in DCIM to remember all the information of the three recipes (first go to the DCIM folder, then just click the \texttt{recipes.jpg} on the top left once to open it, finally make sure you are looking at \texttt{recipes.jpg} and remember the three recipes). Then add the remembered three recipes to Broccoli recipe app. Please make sure the name, description, servings, preparation time, ingredients, and directions are entered correctly. 
& Add the recipes from \texttt{recipes.jpg} in Simple Gallery Pro to the Broccoli recipe app. 
& Reduce target recipes from 3 to 2; Gallery noise from 10 to 1. \\
\midrule

RecipeAddSingleRecipesFromMarkor 
& Ext. 
& Go to Markor and find \texttt{recipes.txt} to remember all the information (name, description, servings, preparation time, ingredients, and directions) of the recipe. Then add the remembered recipe to Broccoli recipe app. Please make sure the name, description, servings, preparation time, ingredients, and directions are entered correctly. 
& Add the recipes from \texttt{recipes.txt} in Markor to the Broccoli recipe app. 
& Reduce Markor noise entries from 10 to 1. \\
\midrule

RecipeAddMultipleRecipesFromMarkor 
& Native 
& Go to Markor and find \texttt{recipes.txt} to remember all the information (name, description, servings, preparation time, ingredients, and directions) of the three recipes. Then add the remembered three recipes to Broccoli recipe app. Please make sure the name, description, servings, preparation time, ingredients, and directions are entered correctly. 
& Add the recipes from \texttt{recipes.txt} in Markor to the Broccoli recipe app. 
& Reduce target recipes from 3 to 2; Markor noise from 10 to 1. \\
\midrule

RecipeAddSingleRecipesFromMarkor2 
& Ext. 
& Go to Markor and find \texttt{recipes.txt} to remember all the information (name, description, servings, ingredients, and directions) of the recipe whose preparation time is \texttt{\{prep\_time\}}. Then add the remembered recipe to Broccoli recipe app. Please make sure the name, description, servings, ingredients, and directions are entered correctly. Please make sure that the preparation time is \texttt{\{prep\_time\}} for the recipe. 
& Add the recipes from \texttt{recipes.txt} in Markor that take \texttt{\{prep\_time\}} to prepare into the Broccoli recipe app. 
& Reduce Markor noise entries from 40 to 1. \\
\midrule

RecipeAddMultipleRecipesFromMarkor2 
& Native 
& Go to Markor and find \texttt{recipes.txt} to remember all the information (name, description, servings, ingredients, and directions) of the recipe whose preparation time is \texttt{\{prep\_time\}}. Then add the remembered recipe to Broccoli recipe app. Please make sure the name, description, servings, ingredients, and directions are entered correctly. Please make sure that the preparation time is \texttt{\{prep\_time\}} for the recipe. 
& Add the recipes from \texttt{recipes.txt} in Markor that take \texttt{\{prep\_time\}} to prepare into the Broccoli recipe app. 
& Reduce target recipes from 3 to 2; Markor noise from 40 to 1. \\
\bottomrule
\end{tabular}
\end{minipage}
}
\caption{Task-level details of AndroidWorld-M. Each difficulty level contains the same 13 task types, while L1 and L2 use more explicit task instructions and L1 additionally introduces controlled simplifications.}
\label{tab:androidworld_m_task_details}
\end{table*}

\begin{table*}[t]
\centering
\scalebox{0.7}{
\begin{minipage}{1.5\textwidth}
\scriptsize
\setlength{\tabcolsep}{3pt}
\renewcommand{\arraystretch}{0.95}
\begin{tabular}{p{3.5cm}p{6cm}p{6cm}p{6cm}}
\toprule
\textbf{Task ID} & \textbf{L1 Task} & \textbf{L2 Task} & \textbf{L3 Task} \\
\midrule

011-SetMeetingAlarmByWorldClock
& Open the Clock app and go to World Clock. Then add `Beijing, China' and `New York, USA' to World Clock. Next, compare the time differences between the two locations and choose any Beijing time that is between 8 AM and 10 PM in both Beijing and New York. Finally, add this Beijing time to your alarm for tomorrow.
& Open the Clock app and go to World Clock. Then add `Beijing, China' and `New York, USA' to World Clock. Next, compare the time differences between the two locations and choose any Beijing time that is between 8 AM and 10 PM in both Beijing and New York. Finally, directly answer this Beijing time.
& Open the Clock app and add `Beijing, China' and `New-York, USA' to the world clock. By comparing their current times, find a suitable meeting time for tomorrow that falls between 8 AM and 10 PM in both cities. Then, set an alarm for this meeting, using the local time in Beijing as the reference. \\
\midrule

012-SetMeetingAlarmByWorldClock
& Open the Clock app and go to World Clock. Then add `Beijing, China' and `Buenos Aires, Argentina' to World Clock. Next, compare the time differences between the two locations and choose any Beijing time that is between 9 AM and 11 PM in both Beijing and Buenos Aires. Finally, add this Beijing time to your alarm for tomorrow.
& Open the Clock app and go to World Clock. Then add `Beijing, China' and `Buenos Aires, Argentina' to World Clock. Next, compare the time differences between the two locations and choose any Beijing time that is between 9 AM and 11 PM in both Beijing and Buenos Aires. Finally, directly answer this Beijing time.
& Open the Clock app and add `Beijing, China' and `Buenos Aires, Argentina' to the world clock. By comparing their current times, find a suitable meeting time for tomorrow that falls between 9 AM and 11 PM in both cities. Then, set an alarm for this meeting, using the local time in Beijing as the reference. \\
\midrule

025-FindAndCompareWikiStats
& Open the Wikipedia app. First, search for `English Wikipedia', scrolling until you find the current number of articles. Remember this number and back to homepage. Then search for `German Wikipedia', scrolling until you find the current number of articles. Remember this number and back to homepage. Then compare the two numbers to see which has more articles. Finally, search again for it (`English Wikipedia' or `German Wikipedia') with more articles and go to its details page.
& Open the Wikipedia app. First, search for `English Wikipedia', scrolling until you find the current number of articles. Remember this number and back to homepage. Then search for `German Wikipedia', scrolling until you find the current number of articles. Remember this number and back to homepage. Then compare the two numbers to see which has more articles. Finally, directly answer which Wikipedia (`English Wikipedia' or `German Wikipedia') has more articles.
& Open the Wikipedia app, search for `English Wikipedia' and find the current number of articles it contains. Remember this number. Then, search for `German Wikipedia' and find its current number of articles. Go to and stay on the page of the Wikipedia edition that has more articles. \\
\midrule

026-FindAndCompareWikiStats
& Open the Wikipedia app. First, search for `Beijing', scrolling until you find the population. Remember this number and back to homepage. Then search for `Shanghai', scrolling until you find the population. Remember this number and back to homepage. Finally, search again for it (`Beijing' or `Shanghai') with the larger population and go to its page.
& Open the Wikipedia app. First, search for `Beijing', scrolling until you find the population. Remember this number and back to homepage. Then search for `Shanghai', scrolling until you find the population. Remember this number and back to homepage. Finally, directly answer which city (`Beijing' or `Shanghai') has larger population.
& Open the Wikipedia app, first search for `Beijing' and remember its population. Next, search for `Shanghai' and remember its population. Compare the two, and then go to and remain on the Wikipedia page for the city with the larger population. \\
\midrule

027-NavigateAndCompareItems
& Open the Wish app. Go to the `Jewelry \& watches' category. First, go to the `Fashion jewelry' subcategory, and then select the cheapest item from the first 6 results. Remember the number of reviews for this item. Next, go to the `Watches' subcategory, and again select the cheapest item from the first 6 results. Remember the number of reviews for this item. Compare the reviews of the two items you remembered. Finally, go back to the product page of the item with more reviews.
& Open the Wish app. Go to the `Jewelry \& watches' category. First, go to the `Fashion jewelry' subcategory, and then select the first item from the first 6 results. Remember the number of reviews for this item. Next, go to the `Watches' subcategory, and again select the first item from the first 6 results. Remember the number of reviews for this item. Compare the reviews of the two items you remembered. Finally, directly answer which item has more reviews.
& In the Wish app, identify two items: the cheapest in `Jewelry \& watches' $\rightarrow$ `Fashion jewelry' (from top 6), and the cheapest in the `Watches' sub-category (from top 6). Remember the number of reviews for both. Navigate to the page of whichever of these two items has more reviews. \\
\midrule

028-NavigateAndCompareItems
& Open the Wish app. Go to the `Office \& tech' category. Then, go to the `Parts \& storage' subcategory and select the product with the most reviews from the first six results. Remember the price of this product. Next, go to the `Hardware' subcategory and select the product with the most reviews from the first six results. Remember the price of this product. Compare the prices of the two products you remembered. Finally, go back to the product page of the cheaper one.
& Open the Wish app. Go to the `Office \& tech' category. Then, go to the `Parts \& storage' subcategory and select the product with the most reviews from the first six results. Remember the price of this product. Next, go to the `Hardware' subcategory and select the product with the most reviews from the first six results. Remember the price of this product. Compare the prices of the two products you remembered. Finally, directly answer which product is cheaper.
& In the Wish app, identify two items: the one with the most reviews in `Office \& tech' $\rightarrow$ `Parts \& storage' (from top 6), and the one with the most reviews in the `Hardware' sub-category (from top 6). Remember the price for both. Navigate to the page of whichever of these two items is cheaper. \\
\midrule

035-CheckProductCompatibility
& Open Amazon app, search for `Anker 737 Power Bank'. Remember its price, capacity (in mAh), and all output port types (make sure all information is found; otherwise, continue searching). Then, back to homepage and search for `MacBook Air M2' and remember its charging port. Next, back to homepage and search for `iPhone 15 Pro' and remember its charging port. Finally open Joplin app, note all the remembered data and answer if the `Anker 737 Power Bank' can simultaneously charge both `MacBook Air M2' and `iPhone 15 Pro'.
& Open Amazon app, search for `Anker 737 Power Bank'. Remember its output port types. Then, back to homepage and search for `MacBook Air M2' and remember its charging port. Next, back to homepage and search for `iPhone 15 Pro' and remember its charging port. Finally, directly answer if the `Anker 737 Power Bank' can simultaneously charge both `MacBook Air M2' and `iPhone 15 Pro'.
& Open Amazon app, search for `Anker 737 Power Bank'. Find its price, capacity (in mAh), and all output port types. Then, search for `MacBook Air M2' and find its required charging port. Next, search for `iPhone 15 Pro' and find its charging port. In Joplin app, note all the collected data and answer if the power bank can simultaneously charge both devices. \\
\midrule

036-CheckProductCompatibility
& Open Amazon app, search for `Samsung T7 Shield SSD'. Remember its USB port specifications, storage capacity, and read/write speeds (make sure all information is found; otherwise, continue searching). Then, back to homepage and search for `PlayStation 5' and remember its USB port specifications. Next, back to homepage and search for `Xbox Series X' and remember its USB port specifications. Finally open Joplin app, note all the remembered data and answer if the `Samsung T7 Shield SSD' is fully compatible with both `PlayStation 5' and `Xbox Series X' consoles' USB standards for external storage.
& Open Amazon app, search for `Samsung T7 Shield SSD'. Remember its USB port specifications. Then, back to homepage and search for `PlayStation 5' and remember its USB port specifications. Next, back to homepage and search for `Xbox Series X' and remember its USB port specifications. Finally, directly answer if the `Samsung T7 Shield SSD' is fully compatible with both `PlayStation 5' and `Xbox Series X' consoles' USB standards for external storage.
& Open Amazon app, search for `Samsung T7 Shield SSD'. Find its price, storage capacity, and read/write speeds. Then search for `PlayStation 5' and find its USB port specifications. Next, search for `Xbox Series X' and find its USB port specifications. In Joplin app, note all collected data and answer if the SSD is fully compatible with both consoles' USB standards for external storage. \\
\midrule

041-CheckRamCompatibilityAndPrice
& Open the Amazon app, search for ``32GB DDR5 RAM'', filter for the ``Corsair'' brand. Remember the price and clock speed of the first item in the search results. Then, open the Bing app and directly search for ``ASUS ROG Strix Z790-E motherboard maximum supported memory speed'' by typing this text. Finally, confirm if the clock speed of that Corsair 32GB DDR5 RAM is less than or equal to the ASUS ROG Strix Z790-E motherboard's maximum supported speed. Directly answer with its price if it is compatible, or ``Not compatible'' if it isn't.
& Open the Amazon app, search for ``32GB DDR5 RAM'', filter for the ``Corsair'' brand. Remember the clock speed of the first item in the search results. Then, open the Bing app and directly search for ``ASUS ROG Strix Z790-E motherboard maximum supported memory speed'' by typing this text. Finally, confirm if the clock speed of that Corsair 32GB DDR5 RAM is less than or equal to the ASUS ROG Strix Z790-E motherboard's maximum supported speed. Directly answer with its price if it is compatible, or ``Not compatible'' if it isn't.
& Open the Amazon app, search for ``32GB DDR5 RAM'', filter for the ``Corsair'' brand, and remember the price and clock speed of the first item in the search results. Then, open the Bing app and search for ``ASUS ROG Strix Z790-E motherboard maximum supported memory speed''. Finally, confirm if the clock speed of that Corsair RAM is less than or equal to the motherboard's maximum supported speed. Directly answer with its price if it is compatible, or ``Not compatible'' if it isn't. \\
\midrule

042-CheckSsdCompatibilityAndPrice
& Open the Amazon app, search for ``1TB External SSD'', filter for the ``Samsung'' brand. Remember the clock speed of the first item in the search results. Then, open the Bing app and directly search for ``PlayStation 5 external SSD speed requirement'' by typing this text. Finally, confirm if the Samsung 1TB External SSD's speed meets or exceeds the PlayStation 5's requirement. Directly answer with its price if it is compatible, or ``Not compatible'' if it isn't.
& Open the Amazon app, search for ``1TB External SSD'', filter for the ``Samsung'' brand. Remember the clock speed of the first item in the search results. Then, open the Bing app and directly search for ``PlayStation 5 external SSD speed requirement'' by typing this text. Finally, confirm if the Samsung 1TB External SSD's speed meets or exceeds the PlayStation 5's requirement. Directly answer with its price if it is compatible, or ``Not compatible'' if it isn't.
& Open the Amazon app, search for ``1TB External SSD'', filter for the ``Samsung'' brand, and remember the price and the read/write speed of the first item in the search results. Then, open the Bing app and search for ``PlayStation 5 external SSD speed requirement''. Finally, confirm if the SSD's speed meets or exceeds the PS5's requirement. Directly answer with its price if it is compatible, or ``Not compatible'' if it isn't. \\
\midrule

063-SearchImageAndDescribe
& Open the bing app, switch to ``image'', and then perform an image search for `Global Smartphone Shipments Market Share 2021'. From the image results, find the image with the chart that specifically displays the data for Q3 2021. Click this image to open it, then identify the top three brands from this Q3 chart and remember their names and their exact market share percentages. Finally, open the joplin app, create a new note titled `Smartphone Market Share 2021 Q3', and write the top three brands with their corresponding percentages.
& Open the bing app, switch to ``image'', and then perform an image search for `Global Smartphone Shipments Market Share 2021'. From the image results, find the image with the chart that specifically displays the data for Q3 2021. Click this image to open it, then identify the top three brands from this Q3 chart and remember their names and their exact market share percentages. Finally, directly answer the top three brands with their corresponding percentages.
& Open the bing app and perform an image search for `Global Smartphone Shipments Market Share 2021'. From the image results, locate and carefully analyze the chart that specifically displays the data for Q3 2021. Identify the top three brands from this Q3 chart, and remember their names and their exact market share percentages. Finally, open the joplin app, create a new note titled `Smartphone Market Share 2021 Q3', and list the top three brands with their corresponding percentages. \\
\midrule

064-SearchImageAndCreateNote
& Open the bing app, switch to ``image'', and then perform an image search for `Global Vehicle Sales Trend by region November 2023'. From the image results, find the first image with clear chart. Click this image to open it, then identify the top three regions with the highest sales growth or volume and remember names of these regions and their corresponding data values (e.g., sales numbers or percentage growth). Finally, open the joplin app, create a new note titled `Vehicle Sales Trend November 2023', and write the top three regions with their data.
& Open the bing app, switch to ``image'', and then perform an image search for `Global Vehicle Sales Trend by region November 2023'. From the image results, find the first image with clear chart. Click this image to open it, then identify the top three regions with the highest sales growth or volume and remember names of these regions and their corresponding data values (e.g., sales numbers or percentage growth). Finally, directly answer the top three regions with their data.
& Open the bing app and perform an image search for `Global Vehicle Sales Trend by region November 2023'. Carefully analyze the first clear chart that appears in the search results. From this chart, identify the top three regions with the highest sales growth or volume, and remember the names of these regions and their corresponding data values (e.g., sales numbers or percentage growth). Finally, open the joplin app, create a new note titled `Vehicle Sales Trend November 2023', and list the top three regions with their data. \\
\bottomrule
\end{tabular}
\end{minipage}
}
\caption{Task descriptions of MemGUI-Bench under the three-level protocol (Part I).}
\label{tab:memgui_bench_details_part1}
\end{table*}

\begin{table*}[t]
\centering
\scalebox{0.7}{
\begin{minipage}{1.5\textwidth}
\scriptsize
\setlength{\tabcolsep}{3pt}
\renewcommand{\arraystretch}{0.95}
\begin{tabular}{p{3.5cm}p{6cm}p{6cm}p{6cm}}
\toprule
\multicolumn{4}{l}{\textit{Table~\ref{tab:memgui_bench_details_part1} continued.}} \\
\midrule
067-PlanAndMessageRoute
& Open the Citymapper app, first search for the journey from the White House to the Lincoln Memorial, and remember the fastest one among Public Transit, Walking, and Taxi. Then search for the journey from the Lincoln Memorial to the National Air and Space Museum, and remember the fastest one among Public Transit, Walking, and Taxi. Determine the fastest possible total travel time by combining the modes for two leg. Finally, send a message to +8613811118888 stating this fastest route combination and the total time.
& Open the Citymapper app, first search for the journey from the White House to the Lincoln Memorial, and remember the fastest one among Public Transit, Walking, and Taxi. Then search for the journey from the Lincoln Memorial to the National Air and Space Museum, and remember the fastest one among Public Transit, Walking, and Taxi. Determine the fastest possible total travel time by combining the modes for two leg. Finally, directly answer the fastest route combination and the total time.
& In the Citymapper app, find the travel times for a two-leg journey in Washington D.C.: 1) The White House to the Lincoln Memorial, and 2) the Lincoln Memorial to the National Air and Space Museum. For each leg, find the times for Public Transit, Walking, and Taxi. Determine the fastest possible total travel time by combining the modes for each leg. Send a message to +8613811118888 stating this fastest route combination and the total time. \\
\midrule

068-PlanAndMessageRoute
& Open the Citymapper app, first search for the journey from the Statue of Liberty to the Empire State Building, and remember the fastest one among Public Transit, Walking, and Taxi. Then search for the journey from the Empire State Building to the Metropolitan Museum of Art, and remember the fastest one among Public Transit, Walking, and Taxi. Determine the fastest possible total travel time by combining the modes for two leg. Finally, send a message to +8613822229999 stating this fastest route combination and the total time.
& Open the Citymapper app, first search for the journey from the Statue of Liberty to the Empire State Building, and remember the fastest one among Public Transit, Walking, and Taxi. Then search for the journey from the Empire State Building to the Metropolitan Museum of Art, and remember the fastest one among Public Transit, Walking, and Taxi. Determine the fastest possible total travel time by combining the modes for two leg. Finally, directly answer the fastest route combination and the total time.
& In the Citymapper app, find the travel times for a two-leg journey in New York City: 1) Statue of Liberty to the Empire State Building, and 2) the Empire State Building to The Metropolitan Museum of Art. For each leg, find the times for Public Transit, Walking, and Taxi. Determine the fastest possible total travel time by combining the modes for each leg. Send a message to +8613822229999 stating this fastest route combination and the total time. \\
\midrule

077-CheckAndNotePermissions
& Open the Setting app. Search for `Special app access' and open it. First, view the apps with `Wi-Fi control' permission and remember the list of app names. Next, back to previous page and continue to view the apps allowed `Picture-in-picture' access and remember that list of names. Finally, open the joplin app, create a note titled `App Access Permissions', and write the names you remembered under two headings: `Wi-Fi Control' and `Picture-in-picture'.
& Open the Setting app. Search for `Special app access' and open it. First, view the apps with `Wi-Fi control' permission and remember the list of app names. Next, back to previous page and continue to view the apps allowed `Picture-in-picture' access and remember that list of names. Finally, directly answer the names you remembered.
& Open the Setting app and go to `Special app access'. First, view the apps with `Wi-Fi control' permission and remember the list of app names. Next, view the apps allowed `Picture-in-picture' access and remember that list of names. Finally, open the joplin app, create a note titled `App Access Permissions', and list the names you remembered under two headings: `Wi-Fi Control' and `Picture-in-picture'. \\
\midrule

078-CheckAndNotePermissions
& Open the Setting app. Search for `Special app access' and open it. First, check how many apps have `Wi-Fi control' permission and remember the count. Next, check how many apps are allowed to `Install unknown apps' and remember that count. Finally, open the joplin app, create a note titled `App Permissions' that records both counts.
& Open the Setting app. Search for `Special app access' and open it. First, check how many apps have `Wi-Fi control' permission and remember the count. Next, check how many apps are allowed to `Install unknown apps' and remember that count. Finally, directly answer the counts you remembered.
& Open the Setting app and go to `Special app access'. First, check how many apps have `Wi-Fi control' permission and remember the count. Then, check how many apps are allowed to `Install unknown apps' and remember that count. Finally, open the joplin app and create a note titled `App Permissions' that records both counts. \\
\midrule

093-MultiAppPriceComparison
& Open the Amazon app and search for `Bose QuietComfort Ultra Headphones'. Remember the price of it. Then open the Wish app and search for `Bose QuietComfort Ultra Headphones'. Remember the price of it. The currencies are different, you need to open the bing app to search for the exchange rate. Finally answer with the name of the app, `Amazon' or `Wish', where the price is lower.
& Open the Amazon app and search for `Bose QuietComfort Ultra Headphones'. Remember the price of it. Then open the Wish app and search for `Bose QuietComfort Ultra Headphones'. Remember the price of it. Finally, directly answer the two price.
& Search for `Bose QuietComfort Ultra Headphones' on both the Amazon and Wish apps, remembering the price and currency from each. If the currencies are different, use the bing app to find the exchange rate to compare them. Directly answer with the name of the app, `Amazon' or `Wish', where the price is lower. \\
\midrule

094-MultiAppPriceComparison
& Open the Amazon app and search for `iPhone 16 Pro Max'. Remember the price of it. Then open the Wish app and search for `Bose QuietComfort Ultra Headphones'. Remember the price of it. The currencies are different, you need to open the bing app to search for the exchange rate. Finally answer with the name of the app, `Amazon' or `Wish', where the price is lower.
& Open the Amazon app and search for `iPhone 16 Pro Max'. Remember the price of it. Then open the Wish app and search for `Bose QuietComfort Ultra Headphones'. Remember the price of it. The currencies are different, you need to open the bing app to search for the exchange rate. Finally, directly answer the two price.
& Search for `iPhone 16 Pro Max' on both the Amazon and Wish apps, remembering the price and currency from each. If the currencies are different, use the bing app to find the exchange rate to compare them. Directly answer with the name of the app, `Amazon' or `Wish', where the price is lower. \\
\midrule

104-ResearchAndNoteF1Principals
& Open the BBC Sports app, go to the `Formula 1' $\rightarrow$ `Constructors'' standings. Remember the names and points of the top three teams. Then open the Bing app. For each team, search for their ``team principal'' name. Finally open the Joplin app, create a note titled ``F1 Top 3 Principals'', and write each team with their points and team principal.
& Open the BBC Sports app, go to the `Formula 1' $\rightarrow$ `Constructors'' standings. Remember the names and points of the top three teams. Then open the Bing app. For each team, search for their ``team principal'' name. Finally, directly answer each team with their points and team principal.
& In BBC Sports app, go to the `Formula 1' $\rightarrow$ `Constructors'' standings. Remember the names and points of the top three teams. Then, for each team, search in Bing app for their ``team principal'' name. In Joplin app, create a note titled ``F1 Top 3 Principals'' listing each team with their points and team principal. \\
\midrule

109-AnalyzeMarketAndInvest
& Open the bing app, search for `NVIDIA stock price', and remember the current stock price. Then open the AP News app, search for `NVIDIA earnings report' and go to the first news (not NVIDIA corp), then remember the reported revenue growth rate. Finally, open the Calculator app, input: stock price * (1 + revenue growth rate), and then calculate the result.
& Open the bing app, search for `NVIDIA stock price', and remember the current stock price. Then open the AP News app, search for `NVIDIA earnings report' and go to the first news (not NVIDIA corp), then remember the reported revenue growth rate. Finally, directly answer the stock price and revenue growth rate.
& Open the bing app, search for `NVIDIA stock price', and remember the current price. Then open the AP News app, find the latest news about NVIDIA earnings report, and remember the reported revenue growth rate. Finally, open the Calculator app and calculate: stock price * (1 + revenue growth rate). \\
\midrule

110-AnalyzeMarketAndInvest
& Open the bing app, search for `Apple stock price', and remember the current stock price. Then open the AP News app, search for `Apple earnings report' and go to the first news (not Apple corp), then remember the reported revenue growth rate. Finally, open the Calculator app, input: stock price * (1 + revenue growth rate), and then calculate the result.
& Open the bing app, search for `Apple stock price', and remember the current stock price. Then open the AP News app, search for `Apple earnings report' and go to the first news (not Apple corp), then remember the reported revenue growth rate. Finally, directly answer the stock price and revenue growth rate.
& Open the bing app, search for `Apple stock price', and remember the current price. Then open the AP News app, find the latest news about Apple earnings report, and remember the reported revenue growth rate. Finally, open the Calculator app and calculate: stock price * (1 + revenue growth rate). \\
\midrule

111-SearchAndSetWallpaper
& Open the bing app, switch to ``image'', and then perform an image search for `minimalist black and white wallpaper'. From the results, find the first image that has a portrait (vertical) orientation suitable for a phone wallpaper. Long-press this image to download it. Then, open the Setting app, navigate to wallpaper settings, and change your home screen wallpaper to the image you just downloaded.
& Open the bing app, switch to ``image'', and then perform an image search for `minimalist black and white wallpaper'. From the results, find the first image that has a portrait (vertical) orientation suitable for a phone wallpaper. Long-press this image to download it. Then, open the Setting app, navigate to wallpaper settings, and change your home screen wallpaper to the image you just downloaded.
& Open the bing app and search for `minimalist black and white wallpaper' images. From the results, find the first image that has a portrait (vertical) orientation suitable for a phone wallpaper. Long-press to download this image. Then, open the Setting app, navigate to wallpaper settings, and change your home screen wallpaper to the image you just downloaded. \\
\midrule

112-SearchAndSetWallpaper
& Open the bing app, switch to ``image'', and then perform an image search for `abstract blue ocean wallpaper'. From the results, find the first image that has a portrait (vertical) orientation suitable for a phone wallpaper. Long-press this image to download it. Then, open the Setting app, navigate to wallpaper settings, and change your home screen wallpaper to the image you just downloaded.
& Open the bing app, switch to ``image'', and then perform an image search for `abstract blue ocean wallpaper'. From the results, find the first image that has a portrait (vertical) orientation suitable for a phone wallpaper. Long-press this image to download it. Then, open the Setting app, navigate to wallpaper settings, and change your home screen wallpaper to the image you just downloaded.
& Open the bing app and search for `abstract blue ocean wallpaper' images. From the results, find the first image that is in a portrait (vertical) format. Long-press and download this image. Afterwards, open the Setting app, navigate to the wallpaper settings, and change your lock screen wallpaper to the downloaded image. \\
\midrule

115-FindRecipeAndCalculateCost
& Open the wikiHow app and search for ``how to bake chocolate chip cookies''. Remember the first four ingredients of it. Then open the joplin app and create a checklist named ``Cookie Ingredients'' with the first four ingredients listed. Finally, open the Calculator app and calculate the total cost, assuming each of the four ingredients costs \$3.50.
& Open the wikiHow app and search for ``how to bake chocolate chip cookies''. Remember the first four ingredients of it. Then open the joplin app and create a checklist named ``Cookie Ingredients'' with the first four ingredients listed.
& Open the wikiHow app and search for ``how to bake chocolate chip cookies''. Create a checklist in the joplin app named ``Cookie Ingredients'' with the first four ingredients listed. Then, open the Calculator app and calculate the total cost, assuming each of the four ingredients costs \$3.50. \\
\midrule

116-FindRecipeAndCalculateCost
& Open the wikiHow app and search for ``how to make a pizza''. Remember the first four ingredients of it. Then open the joplin app and create a checklist named ``Pizza Ingredients'' with the first four ingredients listed. Finally, open the Calculator app and calculate the total cost, assuming each of the four ingredients costs \$4.25.
& Open the wikiHow app and search for ``how to make a pizza''. Remember the first four ingredients of it. Then open the joplin app and create a checklist named ``Pizza Ingredients'' with the first four ingredients listed.
& Open the wikiHow app and search for ``how to make a pizza''. Create a checklist in the joplin app named ``Pizza Ingredients'' with the first four ingredients listed. Then, open the Calculator app and calculate the total cost, assuming each of the four ingredients costs \$4.25. \\
\midrule

127-CompareCarsAndResearch
& Open Cars.co.za app, search for a `Volkswagen Polo', and remember the price of the first result. Then back to homepage and search for a `Hyundai i20' and remember the price of the first result. Open the joplin app to write a note that which car is more expensive and by how much. Finally, open the bing app and search for ``Volkswagen Polo vs Hyundai i20 safety rating''. Stay on the search results page.
& Open Cars.co.za app, search for a `Volkswagen Polo', and remember the price of the first result. Then back to homepage and search for a `Hyundai i20' and remember the price of the first result. Finally, directly answer that which car is more expensive and by how much.
& Open Cars.co.za app, search for a `Volkswagen Polo', and remember the price of the first result. Then search for a `Hyundai i20' and remember the price of the first result. Open the Calculator app and find the price difference. Open the joplin app to note which car is more expensive and by how much. Finally, open the bing app and search for ``Volkswagen Polo vs Hyundai i20 safety rating''. Stay on the search results page. \\
\midrule

128-CompareCarsAndResearch
& Open Cars.co.za app, search for a `Ford Ranger', and remember the price of the first result. Then back to homepage and search for a `Toyota Hilux' and remember the price of the first result. Open the joplin app to write a note that which car is more expensive and by how much. Finally, open the bing app and search for ``Ford Ranger vs Toyota Hilux reliability''. Stay on the search results page.
& Open Cars.co.za app, search for a `Ford Ranger', and remember the price of the first result. Then back to homepage and search for a `Toyota Hilux' and remember the price of the first result. Finally, directly answer that which car is more expensive and by how much.
& Open Cars.co.za app, search for a `Ford Ranger', and remember the price of the first result. Then search for a `Toyota Hilux' and remember the price of the first result. Open the Calculator app and find the price difference. Open the joplin app to note which car is more expensive and by how much. Finally, open the bing app and search for ``Ford Ranger vs Toyota Hilux reliability''. Stay on the search results page. \\
\bottomrule
\end{tabular}
\end{minipage}
}
\caption{Task descriptions of MemGUI-Bench under the three-level protocol (Part II, continued from Table~\ref{tab:memgui_bench_details_part1}).}
\label{tab:memgui_bench_details_part2}
\end{table*}

\begin{table*}[t]
\centering
\small
\renewcommand{\arraystretch}{1}
\begin{tabular}{p{0.96\textwidth}}
\toprule
\textbf{System Prompt} \\
\midrule
\ttfamily
\# Tools \\
\\
You may call one or more functions to assist with the user query. \\
\\
You are provided with function signatures within <tools></tools> XML tags: \\
<tools> \\
\{"type": "function", "function": \{"name\_for\_human": "mobile\_use", "name": "mobile\_use", "description": "Use a touchscreen to interact with a mobile device, and take screenshots. \\
* This is an interface to a mobile device with touchscreen. You can perform actions like clicking, typing, swiping, etc. \\
* Some applications may take time to start or process actions, so you may need to wait and take successive screenshots to see the results of your actions. \\
* The screen's resolution is 1000x1000. \\
* Make sure to click any buttons, links, icons, etc with the cursor tip in the center of the element. Don't click boxes on their edges unless asked.", "parameters": \{"properties": \{"action": \{"description": "The action to perform. The available actions are: \\
* \textasciigrave key\textasciigrave : Perform a key event on the mobile device. \\
    - This supports adb's \textasciigrave keyevent\textasciigrave  syntax. \\
    - Examples: \textbackslash "volume\_up\textbackslash ", \textbackslash "volume\_down\textbackslash ", \textbackslash "power\textbackslash ", \textbackslash "camera\textbackslash ", \textbackslash "clear\textbackslash ". \\
* \textasciigrave click\textasciigrave : Click the point on the screen with coordinate (x, y). \\
* \textasciigrave long\_press\textasciigrave : Press the point on the screen with coordinate (x, y) for specified seconds. \\
* \textasciigrave swipe\textasciigrave : Swipe from the starting point with coordinate (x, y) to the end point with coordinates2 (x2, y2). \\
* \textasciigrave type\textasciigrave : Input the specified text into the activated input box. \\
* \textasciigrave system\_button\textasciigrave : Press the system button. \\
* \textasciigrave open\textasciigrave : Open an app on the device. \\
* \textasciigrave wait\textasciigrave : Wait specified seconds for the change to happen. \\
* \textasciigrave answer\textasciigrave : Terminate the current task and output the answer. \\
* \textasciigrave interact\textasciigrave : Resolve the blocking window by interacting with the user. \\
* \textasciigrave terminate\textasciigrave : Terminate the current task and report its completion status.", "enum": ["key", "click", "long\_press", "swipe", "type", "system\_button", "open", "wait", "answer", "interact", "terminate"], "type": "string"\}, \\
"coordinate": \{"description": "(x, y): The x (pixels from the left edge) and y (pixels from the top edge) coordinates to move the mouse to. Required only by \textasciigrave action=click\textasciigrave , \textasciigrave action=long\_press\textasciigrave , and \textasciigrave action=swipe\textasciigrave .", "type": "array"\}, \\
"coordinate2": \{"description": "(x, y): The x (pixels from the left edge) and y (pixels from the top edge) coordinates to move the mouse to. Required only by \textasciigrave action=swipe\textasciigrave .", "type": "array"\}, \\
"text": \{"description": "Required only by \textasciigrave action=key\textasciigrave , \textasciigrave action=type\textasciigrave , \textasciigrave action=open\textasciigrave , \textasciigrave action=answer\textasciigrave , and \textasciigrave action=interact\textasciigrave .", "type": "string"\}, \\
"time": \{"description": "The seconds to wait. Required only by \textasciigrave action=long\_press\textasciigrave  and \textasciigrave action=wait\textasciigrave .", "type": "number"\}, \\
"button": \{"description": "Back means returning to the previous interface, Home means returning to the desktop, Menu means opening the application background menu, and Enter means pressing the enter. Required only by \textasciigrave action=system\_button\textasciigrave ", "enum": ["Back", "Home", "Menu", "Enter"], "type": "string"\}, \\
"status": \{"description": "The status of the task. Required only by \textasciigrave action=terminate\textasciigrave .", "type": "string", "enum": ["success", "failure"]\}\}, \\
"required": ["action"], "type": "object"\}, "args\_format": "Format the arguments as a JSON object."\}\} \\
</tools> \\
\\
For each function call, return a json object with function name and arguments within <tool\_call></tool\_call> XML tags: \\
<tool\_call> \\
\{"name": <function-name>, "arguments": <args-json-object>\} \\
</tool\_call> \\
\\
\# Response format \\
\\
Response format for every step: \\
1) Action: a short imperative describing what to do in the UI. \\
2) Memory: the key information need to memory in the screenshot. \\
3) A single <tool\_call>...</tool\_call> block containing only the JSON: \{"name": <function-name>, "arguments": <args-json-object>\}. \\
\\
Rules: \\
- Output exactly in the order: Action, Memory, <tool\_call>. \\
- Be brief: one for Action. \\
- Do not output anything else outside those three parts. \\
- If finishing, use action=terminate in the tool call. \\
\\
\bottomrule
\end{tabular}
\caption{System prompt used by Stamp-GUI.}
\label{tab:system_prompt}
\end{table*}

\begin{table*}[t]
\centering
\small
\renewcommand{\arraystretch}{1}
\begin{tabular}{p{0.96\textwidth}}
\toprule
\textbf{Dynamic Testing Planner Prompt} \\
\midrule
\ttfamily
I have a mobile operation task: \\
\{self.task\} \\
\\
The detailed task guideline is: \\
\{self.guideline\} \\
\\
The calculation part may be difficult, so I am directly giving you some hints: \\
\{self.calculation\} \\
\\
The final answer to input is: \\
\{self.answer\} \\
\\
Now the phone has entered the home page of the task app, as shown in the image below. \\
\\
Please output exactly one next action. You may choose one of the following four actions: \\
Click - Please describe the element to click from the perspectives of position, shape, function, or other visual characteristics. \\
Input - Please describe the element to type into, and also provide the text content to input. This action will first click the target element to activate the input box and then type the text. \\
Scroll - If you need to scroll down to see more content below, use this action. No extra description is needed. \\
End - If you have entered the correct answer and seen ``Success!'', you may use this action to stop. No extra description is needed. \\
\\
Please use the following JSON format in your reply: \\
```json \\
\{\{"think": "Your detailed reasoning about what to do next, based on the action history, past and current state, task requirement, guideline, and other relevant information", "action": "The next action. Please output it strictly according to the format and requirements above"\}\} \\
``` \\
\\
Please note: \\
1. Follow the task guideline strictly. Do not perform exploratory actions based on your own assumptions. \\
2. The input action already includes one click, so you do not need to click the input box separately before inputting. \\
3. If a popup appears, try closing it first. \\
4. Do not try to input the final answer directly. You must first browse all information according to the detailed guideline, and only then input the answer. \\
5. Before entering the final answer, you must review the action history and screenshots to ensure that all required information for the task has already appeared. \\

\bottomrule
\end{tabular}
\caption{Prompt used by the dynamic testing planner.}
\label{tab:prompt_dynamic_planner}
\end{table*}

\begin{table*}[!ht]
\centering
\small
\renewcommand{\arraystretch}{1}
\begin{tabular}{p{0.96\textwidth}}
\toprule
\textbf{Memory-Acc Evaluation Prompt} \\
\midrule
\ttfamily
I have a task to perform using Chrome on my phone, as follows: \\
\{\} \\
Please note that this task requires memorizing specific information to complete. \\
\\
I will now provide the content that needs to be memorized for this task, as follows: \\
\{\} \\
\\
I have a mobile AI agent that has completed this task, and it has memorized the following: \\
\{\} \\
\\
Now, please determine whether the information the mobile phone agent has memorized is accurate based on the information above: \\
1. If the mobile phone agent's memory completely matches the provided content to be memorized, output ``Complete Match''. \\
2. If the mobile phone agent's memory partially matches the provided content to be memorized, output ``Partial Match''. \\
3. If the mobile phone agent's memory does not match the provided content to be memorized at all, output ``No Match''. \\
\\
``Match'' means that both expressions convey the same meaning. They do not need to be completely identical. As long as the core content to be memorized, such as numbers or strings, is the same, it is acceptable. \\
\\
Now, after careful consideration, please directly output your final judgment. You must choose one of ``Complete Match'', ``Partial Match'', or ``No Match'' and do not output any other content. \\

\bottomrule
\end{tabular}
\caption{Prompt used for memory-accuracy evaluation.}
\label{tab:prompt_memory_acc}
\end{table*}

\begin{table*}[!ht]
\centering
\small
\renewcommand{\arraystretch}{1.1}
\begin{tabular}{p{0.96\textwidth}}
\toprule
\textbf{Scenario Generation Agent Prompt} \\
\midrule
\ttfamily
You are a scenario generator. \\
\\
Your task is to generate a scenario JSON for a simulated mobile web app based on the given spec. \\
\\
Requirements: \\
- The scenario must be diverse, realistic, interactive, and verifiable. \\
- There must be a unique gold answer. \\
- You may only use values from spec.fixed\_values as the source of truth for anything related to the gold answer. \\
- Do not invent any gold-related values outside spec.fixed\_values. \\
- You may expand copywriting, list items, titles, distracting information, and UI page structure details. \\
- Output must be strictly valid JSON. \\
- Output JSON only. No explanation. No markdown. \\
- All content must be in English only. \\
- Do not include any Chinese characters anywhere. \\
\\
Input spec: \\
\{json.dumps(spec, ensure\_ascii=False, indent=2)\} \\
\\
You must output a JSON object with exactly the following structure: \\
\{ \\
  "meta": \{ \\
    "seed": <int>, \\
    "platform\_type": <string>, \\
    "app\_name": <string>, \\
    "theme\_keywords": [<string>...] \\
  \}, \\
  "data": \{ \\
    "entities": [ ... ], \\
    "truth": \{ \\
      "gold": <string>, \\
      "explain": <string>, \\
      "facts": \{ ... \} \\
    \} \\
  \}, \\
  "task": \{ \\
    "question": <string>, \\
    "must\_visit": [<string>...], \\
    "final\_output\_format": <string>, \\
    "gold": <string> \\
  \}, \\
  "ui\_requirements": \{ \\
    "tabs": [ ... ], \\
    "key\_testids": [ ... ], \\
    "distraction\_interactions": [ ... at least 5 items ... ], \\
    "pages": [ \\
      \{ \\
        "id": <string>, \\
        "title": <string>, \\
        "purpose": <string>, \\
        "must\_show\_facts": [<string>...] \\
      \} \\
    ] \\
  \} \\
\} \\
\\
Requirements for the gold field: \\
- gold must be a single-line string that can be matched exactly as a string. \\
- Examples: ``123-45A6789'', ``coding-course;emotional-intelligence'', ``AliceSmith'' \\
- Do not include spaces unless truly required by the format. \\
- task.gold must be exactly identical to data.truth.gold. \\

\bottomrule
\end{tabular}
\caption{Prompt used by the scenario generation agent.}
\label{tab:prompt_scenario_generator}
\end{table*}

\begin{table*}[!ht]
\centering
\small
\renewcommand{\arraystretch}{0.6}
\begin{tabular}{p{0.96\textwidth}}
\toprule
\textbf{Task Generation Agent Prompt} \\
\midrule
\ttfamily
You are a task designer. \\
\\
You will receive a scenario for a simulated mobile web app, including platform information, pages, truth facts, and the gold answer. \\
You must rewrite it into an executable, gradable, and automation-friendly task specification named task\_spec. \\
\\
Input scenario JSON: \\
\{json.dumps(scn, ensure\_ascii=False, indent=2)\} \\
\\
Output requirements: \\
- Output strictly valid JSON only. \\
- No explanation. \\
- No markdown. \\
- The task must be a memory task: \\
  the user must read information across at least 2 different pages, remember it, and then go to a final submission page to submit the answer. \\
- The final answer must be unique. \\
- Grading must use exact string match. \\
- All content must be in English only. \\
- Do not include any Chinese characters anywhere. \\
\\
You must output a task\_spec JSON object with exactly the following structure: \\
\{ \\
  "platform": \{ \\
    "app\_name": <string>, \\
    "style\_keywords": [<string>...], \\
    "tabs": [<string>...] \\
  \}, \\
  "task": \{ \\
    "natural\_language": <string>, \\
    "guideline": <string>, \\
    "must\_visit\_pages": [<page\_id>...], \\
    "memory\_items": [ \\
      \{ \\
        "page\_id": <string>, \\
        "fact\_key": <string>, \\
        "how\_to\_find": "Locator hints for automation, preferably referencing data-testid conventions or page module names" \\
      \} \\
    ], \\
    "final\_action": \{ \\
      "page\_id": <string>, \\
      "type": "input\_and\_submit", \\
      "input\_testid": "answer-input", \\
      "submit\_testid": "answer-submit", \\
      "required\_output\_format": <string>, \\
      "gold": <string> \\
    \}, \\
    "grading": \{ \\
      "method": "exact", \\
      "gold": <string>, \\
      "pass\_regex": <string> \\
    \} \\
  \}, \\
  "ui\_contract": \{ \\
    "required\_testids": [<string>...], \\
    "navigation\_contract": [ \\
      "Example: tab-search/tab-home/tab-orders/tab-me must exist and be clickable for navigation, but must not appear on the submission page" \\
    ] \\
  \}, \\
  "anti\_triviality": \{ \\
    "extra\_interactives": [at least 5 irrelevant but reasonable interactive elements], \\
    "notes": <string> \\
  \} \\
\} \\
\\
Hard constraints: \\
1) task.final\_action.gold must be exactly identical to scenario.data.truth.gold. \\
2) task.grading.gold must be exactly identical to task.final\_action.gold. \\
3) task.grading.pass\_regex must match the gold string, preferably using \^...\$. \\
4) required\_testids must include at least: \\
   - all tabs: tab-* \\
   - search input: search-input (if the scenario has a search page) \\
   - final submission: go-submit-answer, answer-input, answer-submit, result \\
   - fact display containers used by memory\_items, such as fact-xxx / followers-name / price-sku \\
5) task.natural\_language must clearly describe the task flow, such as which pages to visit and what kind of information to look for, but it must not reveal the actual values that need to be remembered. \\
6) task.guideline must be a more detailed version of task.natural\_language, and must describe the full workflow, all memory targets, and the final submission process in detail. \\
7) task.guideline must include the exact location and exact content for every memory item, for example: \\
   ``Find the code 1234 in My > Orders > History'' and \\
   ``Find the event code 5678 in Home > Activities > Popular Events''. \\

\bottomrule
\end{tabular}
\caption{Prompt used by the task generation agent.}
\label{tab:prompt_task_generator}
\end{table*}

\begin{table*}[!ht]
\centering
\small
\renewcommand{\arraystretch}{0.6}
\begin{tabular}{p{0.96\textwidth}}
\toprule
\textbf{Webpage Generation Agent Prompt} \\
\midrule
\ttfamily
You are a web app generator. \\
\\
You must generate a single-file offline-runnable HTML document with inline CSS and JavaScript for a simulated mobile web app test environment. \\
Do not use any external network resources, including images, CDNs, web fonts, scripts, or stylesheets. \\
\\
\{fb\} \\
\\
Scenario JSON: \\
\{json.dumps(scn, ensure\_ascii=False, indent=2)\} \\
\\
Task\_spec JSON: \\
\{json.dumps(task\_spec, ensure\_ascii=False, indent=2)\} \\
\\
You must implement the following: \\
1) A mobile app style bottom tab bar using task\_spec.platform.tabs, with corresponding unique data-testid values in the form tab-*. \\
2) Every testid listed in task\_spec.ui\_contract.required\_testids must exist, be unique, and be interactable when appropriate. \\
3) The navigation behavior described in task\_spec.ui\_contract.navigation\_contract must be implemented. \\
4) The truth facts in scenario.data.truth.facts must be displayed on the appropriate pages exactly as specified. \\
   Do not alter truth values or truth text. \\
5) There must be an entry point with data-testid="go-submit-answer" that leads to a dedicated answer submission page containing: \\
   - data-testid="answer-input" \\
   - data-testid="answer-submit" \\
   - data-testid="result" \\
6) Do not use the <select> element to implement the filter function. If you need to implement the filter function, please use multiple clickable buttons instead. \\
\\
Submission and validation requirements: \\
- The one and only correct answer is: \{gold\} \\
- Make sure no distractor content creates ambiguity. \\
- \{gold\} must remain the unique correct answer under all circumstances. \\
- The answer input area must include a format hint using the wording ``Example: xxx'', where xxx can be any harmless example text. \\
\\
After clicking submit: \\
- result must update in a way that supports automated verification. \\
- If and only if the input exactly equals the gold answer, the page must contain one visible standalone element whose text content is exactly: \\
  Success! \\
  The text must match exactly, including capitalization. \\
  There must be no leading/trailing spaces, no extra punctuation, and no extra newline characters. \\
- If the input is not exactly equal to gold, then the page must not contain any element whose text content is exactly Success!. \\
\\
Submission button requirements: \\
- The button or entry to the submission page must appear prominently in the top-right area. \\
- Do not place it as a floating control near the bottom or side where it may be covered by fixed elements. \\
- Preferably make the submission entry visible from every information-bearing page. \\
\\
Submission page requirements: \\
- Clicking the submission entry must navigate to a completely separate page or view that is isolated from the previous content. \\
- The submission page must remove all task\_spec.ui\_contract.navigation\_contract elements, such as bottom tabs, top navigation, floating controls, or other distracting persistent UI. \\
- The submission page should be kept as clean as possible. \\
- There must be a back button in the top-left corner that returns to the home page. \\
\\
Realism and distraction requirements: \\
- Include at least 5 interactive but task-irrelevant UI elements, such as like buttons, favorite toggles, filters, switches, expandable list items, sort buttons, or menu chips. \\
- These distractions must not block the main task flow. \\
- They must not affect the unique correct answer. \\
\\
Adaptation requirements: \\
- Include a proper meta viewport tag for mobile rendering. \\
- The layout must look like a portrait mobile web app. \\
- Avoid obvious desktop-style page layouts. \\
\\
Language requirements: \\
- The entire HTML, including all visible text, labels, placeholders, buttons, tooltips, hints, examples, alerts, validation messages, titles, tab names, and comments embedded in visible content, must be in English only. \\
- Do not include any Chinese characters anywhere in the HTML. \\
\\
Output format: \\
Output JSON only. No explanation. \\
\{ \\
  "html": "<!doctype html>....</html>", \\
  "answer": \{ \\
    "gold": \{json.dumps(gold, ensure\_ascii=False)\}, \\
    "facts": \{json.dumps(facts, ensure\_ascii=False)\} \\
  \} \\
\} \\

\bottomrule
\end{tabular}
\caption{Prompt used by the webpage generation agent.}
\label{tab:prompt_web_generator}
\end{table*}

\begin{table*}[!ht]
\centering
\small
\renewcommand{\arraystretch}{0.7}
\scalebox{0.9}{
\begin{tabular}{p{1.1\textwidth}}
\toprule
\textbf{Static Inspection Agent Prompt} \\
\midrule
\ttfamily
You are a static semantic inspector and HTML fixer. \\
\\
You will receive: \\
- a scenario containing truth.facts and gold \\
- a task\_spec containing task instructions, workflow, and grading rules \\
- a single-file HTML source code string \\
\\
You must inspect the HTML without rendering it. \\
Your goal is to ensure that a normal user, following the task\_spec workflow and reading the required information from the page, will derive one unique answer, and that answer must be exactly equal to gold. \\
If this is not true, you must directly modify the HTML to fix it. \\
\\
\{fb\} \\
\\
You must output strictly valid JSON only, with exactly the following structure: \\
\{ \\
  "ok": <true/false>, \\
  "predicted\_answer": <string>, \\
  "uniqueness": \{ \\
    "is\_unique": <true/false>, \\
    "ambiguity\_reasons": [<string>...] \\
  \}, \\
  "evidence": \{ \\
    "task\_interpretation": <string>, \\
    "key\_observations": [ \\
      \{ \\
        "page\_or\_section": <string>, \\
        "what": <string>, \\
        "where\_in\_html": <string> \\
      \} \\
    ], \\
    "calculation": <string>, \\
    "key\_notes": [<string>, <string>, ...] \\
  \}, \\
  "fix": \{ \\
    "needed": <true/false>, \\
    "strategy": <string>, \\
    "changes": [ \\
      \{ \\
        "type": <"edit\_text"|"edit\_number"|"move\_block"|"hide\_block"|"rename\_label"|"other">, \\
        "summary": <string>, \\
        "risk": <"low"|"medium"|"high"> \\
      \} \\
    ] \\
  \}, \\
  "html": <string> \\
\} \\
\\
Strict rules: \\
1) predicted\_answer must comply with task\_spec.task.final\_action.required\_output\_format, or at least follow the same exact formatting style as task\_spec.task.final\_action.gold. \\
2) ok=true if and only if: \\
   - predicted\_answer is exactly equal to gold \\
   - uniqueness.is\_unique is true \\
   - the task flow is executable and the required information can actually be found in the HTML without contradiction \\
3) evidence.key\_observations must contain enough information for a third party to verify how predicted\_answer was obtained without running the page. \\
4) Use the smallest possible fix strategy, with the following priority: \\
   (a) first adjust distractor values, ordering, labels, or section membership so they no longer affect the task result \\
   (b) second move distractors out of the relevant comparison or calculation area, such as from a ranking list into a recommendation or ad section \\
   (c) avoid changing scenario.truth.facts items whenever possible; if absolutely necessary, mark the change as risk=high and explain why \\
5) For common bugs involving highest value, top-k, sorting, filtering, or ranking, you must check all candidate items that look comparable, including newly introduced distractors, and ensure they do not change the result away from gold. \\
6) You may modify the HTML to make the task interpretation clearer, for example by renaming a section title to ``Comparison scope for this task: Popularity ranking'', but do not leak the answer in instructional text. \\
7) evidence.key\_notes must list all minimal memory units required for the final answer. If the final answer is formed by combining multiple strings, list each minimal component separately. \\
8) All output content must be in English only. \\
9) Do not include any Chinese characters anywhere, including in the repaired HTML. \\
10) Please ensure that the filter functionality of the webpage is not implemented through the <select> element. It must be implemented through multiple clickable buttons. \\
\\
scenario.truth.facts: \\
\{json.dumps(facts, ensure\_ascii=False, indent=2)\} \\
\\
gold: \\
\{gold\} \\
\\
scenario JSON: \\
\{json.dumps(scn, ensure\_ascii=False, indent=2)\} \\
\\
task\_spec JSON: \\
\{json.dumps(task\_spec, ensure\_ascii=False, indent=2)\} \\
\\
HTML source: \\
-----BEGIN HTML----- \\
\{html\} \\
-----END HTML----- \\

\bottomrule
\end{tabular}
}
\caption{Prompt used by the static inspection agent.}
\label{tab:prompt_static_inspector}
\end{table*}

\end{document}